\documentclass[10pt,twocolumn,letterpaper]{article}

\usepackage{cvpr}              % To produce the CAMERA-READY version
\usepackage{bm}
\usepackage{multirow}
\usepackage{makecell}
\usepackage{graphicx}
\usepackage{amsmath}
\usepackage{amssymb}
\usepackage{booktabs}

\usepackage[pagebackref,breaklinks,colorlinks]{hyperref}

\usepackage[capitalize]{cleveref}
\crefname{section}{Sec.}{Secs.}
\Crefname{section}{Section}{Sections}
\Crefname{table}{Table}{Tables}
\crefname{table}{Tab.}{Tabs.}

\def\cvprPaperID{1516} % *** Enter the CVPR Paper ID here
\def\confName{CVPR}
\def\confYear{2022}

\begin{document}

%%%%%%%%% TITLE - PLEASE UPDATE
\title{Weakly Supervised Temporal Action Localization via \\ Representative Snippet Knowledge Propagation}

\author{Linjiang Huang\textsuperscript{\rm 1,2} \quad
        Liang Wang\textsuperscript{\rm 3} \quad
        Hongsheng Li\textsuperscript{\rm 1,2} \thanks{Corresponding author.}\\
\textsuperscript{\rm 1}CUHK-SenseTime Joint Laboratory, The Chinese University of Hong Kong \\
\textsuperscript{\rm 2}Centre for Perceptual and Interactive Intelligence, Hong Kong \\
\textsuperscript{\rm 3}Institute of Automation, Chinese Academy of Sciences \\
{\tt\small ljhuang524@gmail.com, wangliang@nlpr.ia.ac.cn, hsli@ee.cuhk.edu.hk}
}

\maketitle
%%%%%%%%% ABSTRACT
\begin{abstract}
Weakly supervised temporal action localization aims to localize temporal boundaries of actions and simultaneously identify their categories with only video-level category labels. Many existing methods seek to generate pseudo labels for bridging the discrepancy between classification and localization, but usually only make use of limited contextual information for pseudo label generation. To alleviate this problem, we propose a representative snippet summarization and propagation framework. Our method seeks to mine the representative snippets in each video for propagating information between video snippets to generate better pseudo labels. For each video, its own representative snippets and the representative snippets from a memory bank are propagated to update the input features in an intra- and inter-video manner. The pseudo labels are generated from the temporal class activation maps of the updated features to rectify the predictions of the main branch.
Our method obtains superior performance in comparison to the existing methods on two benchmarks, THUMOS14 and ActivityNet1.3, achieving gains as high as 1.2\% in terms of average mAP on THUMOS14. Our code is available at \url{https://github.com/LeonHLJ/RSKP}.
\end{abstract}

%%%%%%%%% BODY TEXT
\section{Introduction} \label{sec:intro}
Temporal action localization in videos has a wide range of applications in different scenarios. This task aims to localize action instances in untrimmed videos along the temporal dimension. Most existing methods \cite{xu2017r,zhao2017temporal,lin2018bsn,chao2018rethinking,lin2019bmn} are trained in a fully supervised manner where both video-level labels and frame-wise annotations are provided. In contrast to these strong supervision based methods, weakly-supervised temporal action localization method attempt to localize action instances in videos, leveraging only video-level supervisions. This setting enables the method to bypass the manual annotations of temporal boundaries, which are laborious, expensive and prone to large variations \cite{schindler2008action}.
\begin{figure}[!t]
  \centering
  \includegraphics[width=0.92\columnwidth]{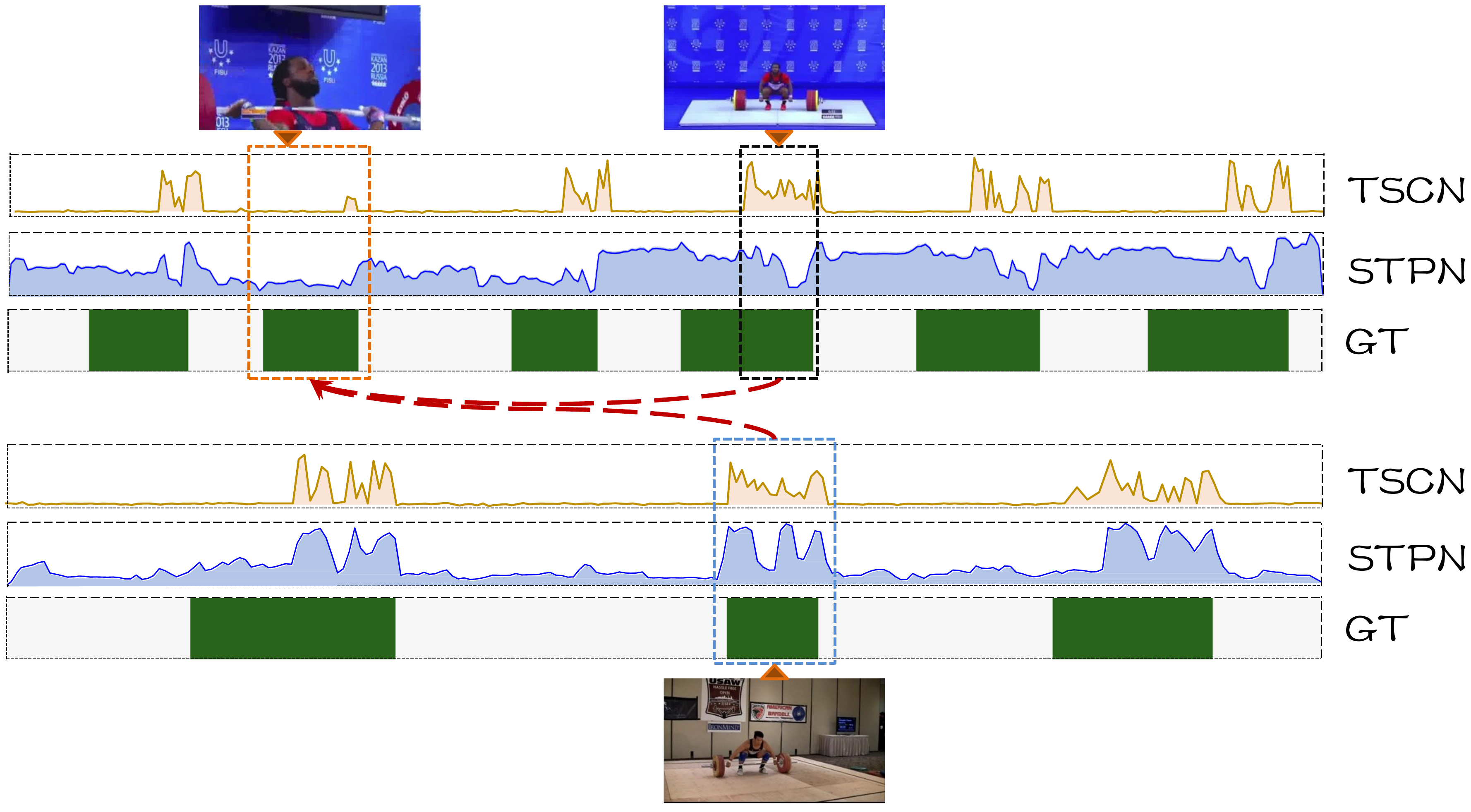}
  \caption{Two examples of the action of ``\emph{Clean and Jerk}''. The barcode is the ground-truth (GT). The line charts are snippet-wise classification scores. We show the detection results of two methods, TSCN \cite{zhai2020two} and STPN \cite{nguyen2018weakly}. The instance in the orange box is hard for both the two methods, because most of the snippets show only a part of the athlete. In contrast, the instances (black box and green box) with frontal view and whole body is much easier to be detected. A direct way to address this issue would be to propagate (red lines) knowledge of representative snippets to other snippets.}\label{fig:motivation}
\vspace{-5mm}
\end{figure}

Due to the absence of fine-grained annotations, existing works mainly embrace a localization-by-classification pipeline \cite{wang2017untrimmednets,zhao2020equivalent}, where a classifier is trained with video-level annotations of action categories \cite{narayan2021d2} and is used to obtain a sequence of class logits or predictions, \ie, temporal class activation maps (TCAMs). Usually, detection results are obtained from TCAMs via a post-processing step \cite{wang2017untrimmednets,nguyen2018weakly} (\eg, thresholding) or a localization branch \cite{liu2019weakly,shou2018autoloc}. Therefore, the quality of TCAMs determines the upper bound of the model. However, there is generally be a discrepancy between classification and localization \cite{liu2019completeness,liu2021weakly,liu2021acsnet}. Since each video generally contains multiple snippets\footnote{In this paper, we view snippets as the smallest granularity since the high-level features of consecutive frames vary smoothly over time \cite{wiskott2002slow,jayaraman2016slow}.}, with only video-level annotations, the model would easily focus on the contextual background or discriminative snippets that contribute most to video-level classification, which hinders the generation of high quality TCAMs.

To address this issue, pseudo label-based methods \cite{pardo2021refineloc,luo2020weakly,zhai2020two,yang2021uncertainty} were proposed to generate snippet-wise pseudo labels for bridging the gap between classification and localization.
However, existing methods only leverage limited information, \ie, the information within each snippet, to generate pseudo labels, which is insufficient to generate high quality pseudo labels.
In Figure \ref{fig:motivation}, we show the detection results of two methods. The first method TSCN \cite{zhai2020two} is a pseudo label-based method, while the second method STPN \cite{nguyen2018weakly} is a simple baseline model without using pseudo labels. As we can see, even if TSCN achieves much gain over STPN, neither of the two methods successfully detects the difficult action instance in the orange box, which only shows partial body of the athlete. Obviously, the pseudo labels generated from the inaccurate TCAMs are also inaccurate. In contrast, for the easy instance, \eg, the one in the blue box, both of the two methods accurately detect it.

The above observation motivate us to introduce contextual information for pseudo label generation. Specifically, we propose to propagate the knowledge of those representative snippets (\eg, the black and blue boxes in Figure \ref{fig:motivation}) in an intra- and inter-video manner to facilitate the pseudo label generation, especially for those difficult snippets (\eg, the orange box in Figure \ref{fig:motivation}). To achieve this goal, a major issue is how to determine the representative snippets and how to propagate their useful knowledge to other snippets.
Besides, it should be effective to summarize and propagate snippet-level knowledge across videos, so as to take advantage of the large variation of videos in a large scale dataset.

We present a representative snippet knowledge propagation framework. To facilitate knowledge propagation, we propose to mine the representative snippets, which can mitigate the influence of outlier snippets to serve as a bridge to propagate knowledge between snippets. Specifically, we utilize the expectation-maximization (EM) attention \cite{li2019expectation} to handle the variations caused by different camera views \cite{zhang2021cola}, sub-action differences \cite{liu2019completeness,huang2021modeling}, confusing background context \cite{liu2021acsnet,liu2021weakly} and to capture the important semantic of each video, which are severed as the representative snippets in our method. After that, we employ a memory bank to store representative snippets of high confidences for each class. This design enables our method to leverage representative snippets in an inter-video fashion, and also avoids much GPU memory cost during training. Furthermore, to propagate the knowledge of representative snippets, we propose a bipartite random walk (BiRW) module, which integrates the random walk operation to update the features of the input video with intra- and inter-video representative snippets. The TCAMs of the updated features serve as online refined pseudo labels to rectify the predictions of the main branch.

The contribution of this paper is three-fold.
(a) We propose a novel representative snippet knowledge propagation framework for weakly supervised temporal action localization, which generates better pseudo labels via representative snippet knowledge propagation to effectively alleviates the discrepancy between classification and detection.
(b) The proposed framework can be applied to most existing methods to consistently improve their localization performance.
(c) Compared with state-of-the-art methods, the proposed framework yields improvements of \textbf{1.2\%} and \textbf{0.6\%} average mAP on THUMOS14 and ActivityNet1.3.
\begin{figure*}[t]
\centering
\includegraphics[width=0.93\textwidth]{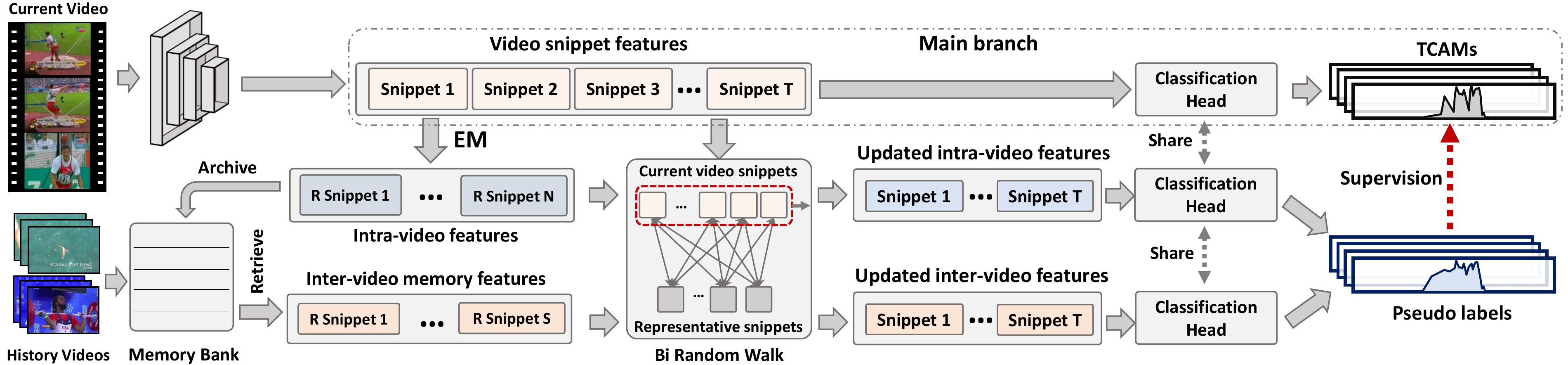}
\caption{The overview of our method. We first extract snippet-wise features by using a fixed-weighted backbone network appended with a small learnable network.
We utilize the expectation-maximization (EM) Attention \cite{li2019expectation} to learn a Gaussian mixture model (GMM) for each video, whose mean vectors are treated as the representative snippets. We use a memory bank to store representative snippets with high prediction scores for each class. To propagate representative snippets, we propose a bipartite random walk (BiRW) module to gradually update the original features. Given the original features and the updated features, we feed them into three parallel classification heads with sharing parameters. The output TCAMs of the two branches of the updated features are first fused and then serve as the online refined pseudo labels to rectify the predictions of the main branch.}
\label{fig:framework}
\vspace{-5mm}
\end{figure*}

\section{Related Work} \label{sec:related_work}

As stated in \cite{liu2019completeness,moniruzzaman2020action,narayan2021d2}, there is generally a discrepancy between classification and localization in this task. Recently, many efforts have been made to solve this issue. We divide these methods into four categories.

The first category is the metric learning-based methods. For example, W-TALC \cite{paul2018w}, 3C-Net \cite{narayan20193c}, RPN \cite{huang2020relational} and A2CL-PT \cite{min2020adversarial} employed the center loss \cite{wen2016discriminative}, clustering loss \cite{yang2018robust}, triplet loss \cite{schroff2015facenet}, \etc to learn intra-class compact features. However, these methods starts from the perspective of classification, that is, they learn video-level foreground features \cite{paul2018w,huang2020relational} or class-wise features \cite{narayan20193c,min2020adversarial} to enforce intra-class compactness. Whereas, these video-level features are aggregated from those discriminative snippets, they can hardly influence those less-discriminative features in the snippet-level.
In contrast, our method is from the perspective of detection, which generates better snippet-wise pseudo labels by propagating the knowledge of representative snippets, so as to obtain better detection results.

The second category is the erasing-based methods, whose representative methods are Step-by-Step Erasion \cite{zhong2018step} and A2CL-PT \cite{min2020adversarial}. These methods is based on the adversarial complementary learning \cite{zhang2018adversarial}, which first finds the discriminative regions and then tries to weight more less-discriminative regions from the remaining regions. However, it is difficult to set a proper number of steps for different categories with different complexities.

The methods of the third category are built on a multi-branch \cite{liu2019completeness} or multi-attention architecture \cite{liu2021weakly,liu2021acsnet,islam2021hybrid,huang2021foreground}. These methods adopted a similar idea to the second category's methods, except for the parallel processing. To avoid trivial solutions, these methods require additional regularization terms to make branches or attention scores different or complementary. Likewise, it is difficult to define a proper number of branches or attentions for all action categories. Our method is also a multi-branch architecture. However, the additional branch generates online pseudo labels for the main branch. Therefore, instead of forcing branches to be different, we force them to be similar.

The fourth category is the pseudo label-based methods. RefineLoc \cite{pardo2021refineloc} is the first method that generates snippet-level hard pseudo labels for WTAL. However, it simply expanded previous detection results to obtain pseudo labels, which may result in over-complete proposals. EM-MIL \cite{luo2020weakly} put the pseudo-label generation into an expectation-maximization framework. TSCN \cite{zhai2020two} proposed to generate snippet-level pseudo labels from late fusion attention sequence. UGCT \cite{yang2021uncertainty} presented an uncertainty guided collaborative training strategy to generate pseudo labels with modality collaborative learning and uncertainty estimation. Despite the success of these methods, they leverage little contextual information for pseudo label generation.

\section{Proposed Method} \label{sec:proposed_method}
In this section, we elaborate on the proposed method. The illustration of the overall method is shown in Figure \ref{fig:framework}.

\vspace{-4mm}
\paragraph{Problem definition.}
Let \(V=\{v_t\}_{t=1}^{L}\) be a video of temporal length \(L\). Each video is divided into a series of non-overlapping snippets. Assume that we have a set of \(N\) training videos \(\{V_i\}_{i=1}^{N}\) being annotated with their action categories \(\{\bm{y}_i\}_{i=1}^{N}\), where \(\bm{y}_i\) is a binary vector indicating the presence/absence of each of $k$ action. During inference, for a video, we predict a set of action instances \(\{(c, q, t_s, t_e)\}\), where \(c\) denotes the predicted action class, \(q\) is the confidence score, \(t_s\) and \(t_e\) represent the start time and end time.

\vspace{-5mm}
\paragraph{Overview.}
We propose a representative snippet knowledge summarization and propagation framework for generating better snippet-level pseudo labels to enhance the final localization performance. Pseudo labels have an important role for bridging the gap between classification and detection. Nevertheless, existing methods merely leverage contextual information for pseudo label generation, leading to inaccurate pseudo labels and compromising performance.
Our key idea is to generate pseudo labels by propagating the knowledge of representative snippets, which act as a bridge between discriminative snippets and less-discriminative snippets, thereby indirectly propagate information between all snippets and producing accurate pseudo labels to improve the model's performance. In our method, we first extract video features by a feature extraction module. After that, we summarize the representative snippets from the extracted video features. The representative snippets with high confidences are maintained in a memory bank. For each video, we leverage both the intra- and inter-video representative snippets that are retrieved from the memory bank. A bipartite random walk module is introduced to update the video features with the two kinds of representative snippets. Given the video features and the updated video features, we feed them into three parallel classification heads with sharing parameters.
The output TCAMs of the two branches corresponding to the updated features are first fused and then serve as the online refined pseudo labels to rectify the predictions of the main branch.

\vspace{-5mm}
\paragraph{Feature extraction.}
Given a video, we first divide it into a series of non-overlapping snippets. Following \cite{nguyen2018weakly,paul2018w}, we utilize a fixed-weight backbone network, Inflated 3D (I3D) \cite{carreira2017quo} pre-trained on the Kinetics-400 dataset \cite{kay2017kinetics}, to encode appearance (RGB) and motion (optical flow) information into a $d=2048$ dimensional feature for each snippet. The I3D features are encoded into latent embeddings \(\bm{{\rm F}} \in \mathbb{R}^{l \times d}\) with a convolutional layer, where $l$ is the number of video snippets of a video. We take \(\bm{{\rm F}}\) as the input of our model.

\vspace{-5mm}
\paragraph{Classification head.}
The classification head is used to generate TCAMs, it can be any existing WSTAL methods. To generate high-quality TCAMs and improve the lower bound of our method, we use the recent  FAC-Net \cite{huang2021foreground} as the classification head for its simple pipeline and promising performance.
Note that, there are some modifications to FAC-Net in our method. First, we discard the class-wise foreground classification head in our method, since the commonly used class-agnostic attention head and multiple instance learning head already enable our method to achieve a high baseline performance. Second, we use the sigmoid rather than the softmax function to obtain normalized foreground scores. This setting enables our method to use the attention normalization term \cite{zhai2020two} to obtain highly confident representative snippets. Third, we do not use the hybrid attention strategy that is designed to alleviate the discrepancy between classification and detection.

In the following sections, we first investigate how to obtain summarizations of the representative snippets and to propagate their information to all other snippets.

\subsection{Representative Snippet Summarization} \label{sec:representative_snippet_summarization}
\begin{figure}[!t]
  \centering
  \includegraphics[width=0.92\columnwidth]{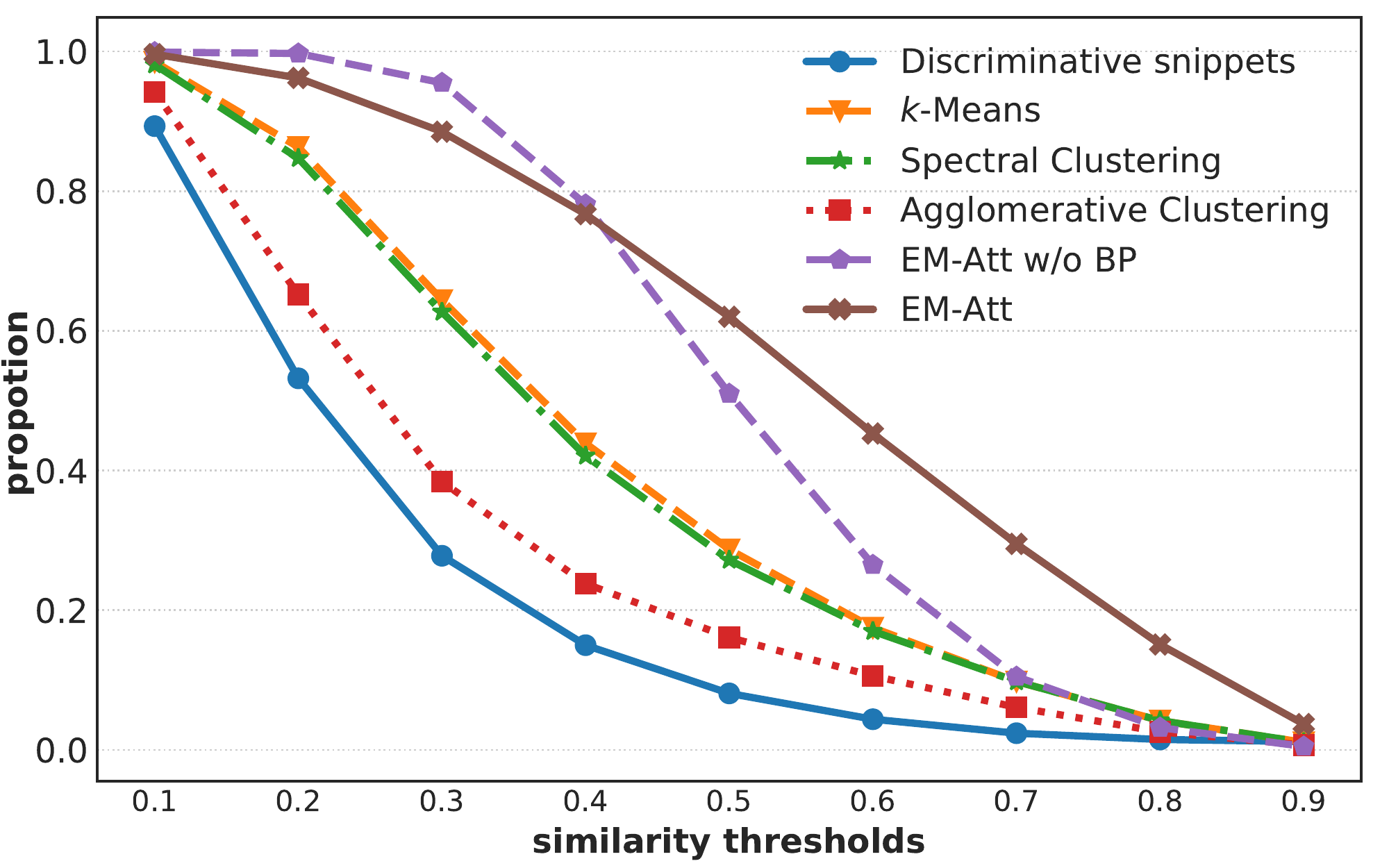}
  \caption{We assign each representative snippet with the label of its nearest snippet and then select the snippets of the same category according to the ground truth. We plot the proportion of the number of snippets, whose similarities to the representative snippet are higher than a certain threshold (0.1:0.9), to the total number of snippets of the same category. The results of EM-Att is calculated upon our full model. Other results are based on the features of a well learned baseline model, which takes the video features $\mathbf{F}$ as input and contains one classification head.}\label{fig:clustering_comparison}
\vspace{-2mm}
\end{figure}

A na\"ive way to obtain representative snippets would be to select the snippets with high prediction scores, \ie, discriminative snippets. However, as shown in Figure \ref{fig:clustering_comparison}, even after large-scale pre-training, there are generally low similarities between discriminative snippets and other snippets of the same category.
Intuitively, representative snippets should be able to describe most of the snippets of the same class, so as to act as a bridge to associate the snippets of the same class for knowledge propagation.
It is therefore ineffective to directly propagate the information of the discriminative snippets to other snippets.
Therefore, we propose to summarize the representations of video snippets to obtain representative snippets of each video. In Figure \ref{fig:clustering_comparison}, we can see that using cluster centers via clustering video snippet features (\eg, $k$-means, spectral clustering and agglomerative clustering) as the representative snippets achieves much better performance on building stronger relations with other snippets. According to our experiments, using the clustering methods to summarize representative snippets is important to achieve high detection performance.

In this work, we employ the expectation-maximization (EM) attention \cite{li2019expectation} to generate the representative snippets of each video. EM attention uses a special EM algorithm based on Gaussian mixture model (GMM) \cite{richardson1997bayesian}.
Specifically, a separated GMM is adopted to capture the feature statistics of each video and models the distribution of \(\mathbf{f}_{i} \in \mathbb{R}^d\) (the $i$-th snippet feature of $\mathbf{F} \in \mathbb{R}^{l \times d}$) as a linear composition of Gaussians as follow,
\begin{equation}
p\left(\mathbf{f}_{i}\right)=\sum\nolimits_{k=1}^{n} z_{ik} \mathcal{N}\left(\mathbf{f}_{i} | \boldsymbol{\mu}_{k}, \boldsymbol{\Sigma}_{k}\right),
\vspace{-1mm}
\end{equation}
where $n$ is the number of Gaussians, $\boldsymbol{\mu}_{k} \in \mathbb{R}^{d}$, $\boldsymbol{\Sigma}_{k} \in \mathbb{R}^{d \times d}$ and $z_{ik}$ denote the mean, covariance and weight for the $k$-th Gaussian. Following \cite{li2019expectation}, we replace the covariance with identity matrix \(\bm{{\rm I}}\) and leave out it in the following equations.

\begin{figure}[!t]
  \centering
  \includegraphics[width=0.92\columnwidth]{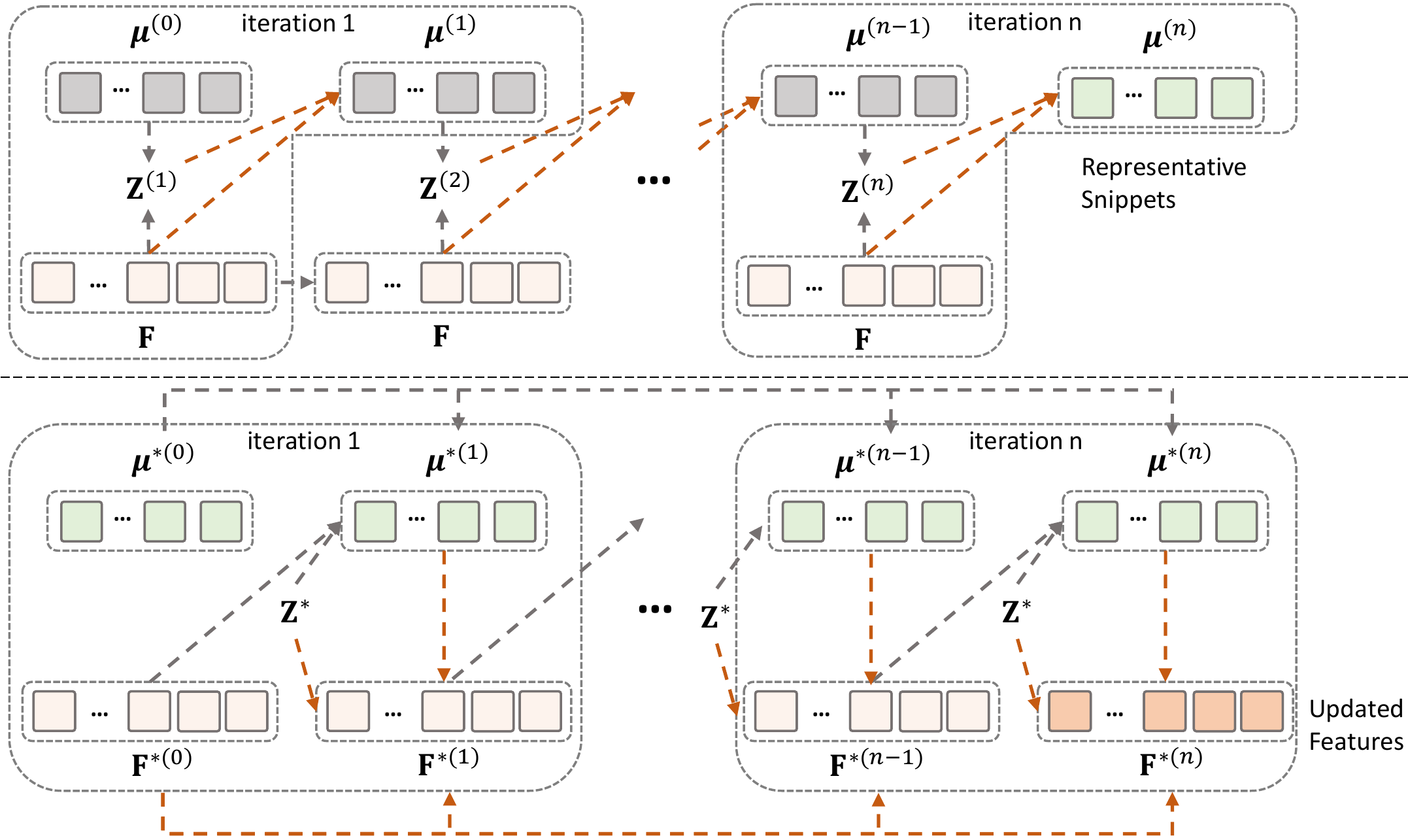}
  \caption{Illustration of the representative snippet summarization (top) and representative snippet propagation (bottom). The gray lines denote the E step, while the red lines denote the M step.}\label{fig:propagation}
\vspace{-6mm}
\end{figure}

As shown in Figure \ref{fig:propagation} (top), the EM attention starts from the randomly initialized means $ \boldsymbol{\mu}^{(0)} \in \mathbb{R}^{n \times d}$. At the $t$-th iteration, it first performs the E step to calculate the new weights $\mathbf{Z}^{(t)} \in \mathbb{R}^{t \times n}$ of Gaussians as
\begin{equation} \label{eq:e_step2}
\mathbf{Z}^{(t)}=\operatorname{softmax}\left(\lambda \operatorname{Norm}_2(\mathbf{F})\operatorname{Norm}_2(\boldsymbol{\mu}^{(t-1)})^{\top}\right),
\vspace{-2mm}
\end{equation}
where $\lambda$ denotes a hyper-parameter to control the smoothness of the distribution. The $\operatorname{Norm}_2(\mathbf{F})$ denotes the $l_2$-norm along each row of $\mathbf{F}$.
%, \ie, $\mathbf{F}=[\mathbf{f}_1^{\top}, \cdots, \mathbf{f}_t^{\top}]^{\top}$, $\operatorname{Norm}_2(\mathbf{F})=[\mathbf{\hat{f}}_1^{\top}, \cdots, \mathbf{\hat{f}}_t^{\top}]^{\top}$, where $\mathbf{\hat{f}}_i = \mathbf{f}_i / \lVert\mathbf{f}_i\rVert_2$
The softmax operation is performed along each row of $\mathbf{Z}$. Therefore, $z_{ik}^{(t)}$ denotes the probability that the snippet feature $\mathbf{f}_i$ is generated by the $k$-th Gaussian.
After the E step, the M step turns to update the means $\boldsymbol{\mu}$ as
\begin{equation} \label{eq:m_step2}
\vspace{-1mm}
\boldsymbol{\mu}^{(t)}= \operatorname{Norm}_1(\mathbf{Z}^{(t)})^{\top} \mathbf{F},
\vspace{-0.5mm}
\end{equation}
where $\operatorname{Norm}_1(\mathbf{Z}^{(t)})$ denotes the column-wise $l_1$ normalization for  $\mathbf{Z}^{(t)}$. We can see that Equation (\ref{eq:m_step2}) updates the means using the weighted summation of the features $\mathbf{F}$. The $i$-th row and the $k$-th column of $\operatorname{Norm}_1(\mathbf{Z}^{(t)})$ represents the membership values of the feature $\mathbf{f}_{i}$ to the $k$-th Gaussian. Besides, the normalization also ensures the updated $\boldsymbol{\mu}$ lies in the same embedding space as $\mathbf{F}$. Therefore, alternately executing Equations (\ref{eq:e_step2}) and (\ref{eq:m_step2}) captures global contexts of a video in a non-local \cite{wang2018non} but more efficient way, owing to the much smaller size of the means $\boldsymbol{\mu}^{(t)} \in \mathbb{R}^{n \times d}$ compared with the video features $\mathbf{F} \in \mathbb{R}^{l \times d}$ ($l \gg n$).

We integrate two EM iterations in our network to obtain the promising representative snippets (\ie, $\boldsymbol{\mu}^{(2)}$). As a result, unlike \cite{li2019expectation} that uses the moving averaging to update the initialized means $\boldsymbol{\mu}^{(0)}$ to prevent gradient vanishing or explosion caused by too many iterations, we update the initialized means by standard back propagation. Besides, it is interesting that when we initialize $\boldsymbol{\mu}^{(0)}$ with a (semi) orthogonal matrix \cite{saxe2013exact}, even if we fix the initialized means (\ie, EM-Att w/o BP in Figure \ref{fig:clustering_comparison}), the obtained representative snippets are more representative than the cluster centers of other clustering methods. In our experiments, EM attention without back propagation achieves much better performance than other clustering methods. When we update the initial means $\boldsymbol{\mu}^{(0)}$ by standard back propagation (\ie, EM-Att in Figure \ref{fig:clustering_comparison}), they can capture the feature distribution of the dataset and achieves the optimal performance.

To make the equations clear and avoid confusion, we denote the online calculated representative snippets as $\boldsymbol{\mu}^{a}$.

\subsection{Representative Snippet Memory Bank}
After obtaining the representative snippets of each video, we use a memory bank to store representative snippets of all videos with high confidences for each class. The insight is that different videos may contain action instances of the same class but with different appearances. Therefore, propagating representative snippets in an inter-video manner via the memory bank takes advantage of the large variation of many videos of each class to facilitate the network to recognize those difficult action instances.

Specifically, we maintain two memory tables for storing the features of representative snippets and their scores, respectively. We denote the memory table of the representative snippets as \(\bm{{\rm M}} \in \mathbb{R}^{c \times s \times d}\), where $c$ is the number of classes and $s$ denotes the number of memory slots (representative snippets) for each class.
Given the representative snippets of a video, we utilize the action classifier in the classification head to obtain their class predictions. We then compare their prediction scores of the ground truth class with the representative snippets in the memory table $\bm{{\rm M}}$, the ones with higher prediction scores are archived into the memory table $\bm{{\rm M}}$.
Simultaneously, the corresponding scores in the score memory table are also updated. In summary, we only keep the representative snippets of the top-$s$ scores in the memory table $\bm{{\rm M}}$. Consequently, for each video, we have the representative snippets that are learned from the current video, and the representative snippets that are retrieved from the memory table $\mathbf{M}$ corresponding to the ground truth classes.

To distinguish from the online representative snippets \(\boldsymbol{\mu}^{a}\), we denote the offline representative snippets as \(\boldsymbol{\mu}^{e}\).

\subsection{Representative Snippet Propagation}
Given the online and offline representative snippets \(\boldsymbol{\mu}^{a}\) and \(\boldsymbol{\mu}^{e}\), a challenge is how to propagate the representative snippets to snippet features $\mathbf{F}$ of the current video. Intuitively, a direct way is to use the affinity $\mathbf{Z}^{*}$ ($*$ is $a$ or $e$ corresponding to \(\boldsymbol{\mu}^{a}\) or \(\boldsymbol{\mu}^{e}\)) calculated by Equation (\ref{eq:e_step2}) to conduct a random walk operation \cite{aldous2002reversible} as $\mathbf{Z}^{*}\boldsymbol{\mu}^{*}$. In practice, we would like the updated features do not deviate too far from the video features $\bm{{\rm F}}$. Therefore, the propagation process can be formulated as $\mathbf{F}^{*} = w \cdot \mathbf{Z}^{*}\boldsymbol{\mu}^{*} + (1-w) \cdot \mathbf{F}$, where $w$ denotes a parameter $[0, 1]$ that controls the trade-off between feature propagation and the original features.
However, even if the representative snippets have high similarities with most of the snippets of the same class, it is impractical to fully propagate the knowledge of the representative snippets via a single pass of propagation.
We find that, the representative snippets $\boldsymbol{\mu}^{*}$ and the video features \(\mathbf{F}\) actually make up a complete bipartite graph, whose affinities are depicted by \(\mathbf{Z}^{*}\). Therefore, we propose a bipartite random walk (BiRW) module to enable multiple passes of propagation to fully fuse representative snippets’ knowledge into the video snippet features.

%A direct way is to utilize the re-estimation of the input features \(\bm{{\rm F}}\), \ie, \(\bm{{\rm F}}^{re}=\bm{{\rm Z}}^{(new)}\boldsymbol{\mu}\), where $\bm{{\rm Z}}^{(new)}$ denotes the weights (Equation (\ref{eq:e_step2})) at the last iteration of the EM process\footnote{For \(\boldsymbol{\mu}^{e}\), we use Equation (\ref{eq:e_step2}) to calculate $\bm{{\rm Z}}^{(new)}$.}. Whereas, we empirically find the features \(\bm{{\rm F}}^{re}\) lost too many details to make the model accurately distinguish actions from background, leading to poor performance.

%To achieve this goal, we propose a bipartite random walk (BiRW) module, which is an extension of the random walk operation \cite{aldous2002reversible} in a bipartite manner, to smoothly update the original features in an iterative way.

%As we can see the EM process in Section \ref{sec:representative_snippet_summarization} fixes the input features \(\bm{{\rm F}}\) and alternately learn new \(\bm{{\rm Z}}\) and \(\boldsymbol{\mu}\).
%After several EM iterations, the \(\boldsymbol{\mu}\) would gradually converge and the corresponding weights \(\bm{{\rm Z}}\) are stable. At this time, the representative snippets and the features \(\bm{{\rm F}}\) make up a complete bipartite graph, whose connections \(\bm{{\rm Z}}\) can be viewed as fixed.
Specifically, there are multiple iterations in the BiRW. At the $t$-th iteration, the propagation process is formulated as
\begin{align}\label{eq:birw1}
  \vspace{-1mm}
  \boldsymbol{\mu}^{*(t)} =& w \cdot \operatorname{Norm}_1(\mathbf{Z}^{*})^{\top} \mathbf{F}^{*(t-1)} + (1-w) \cdot \boldsymbol{\mu}^{*(0)}, \\
  \mathbf{F}^{*(t)} =& w \cdot \mathbf{Z}^{*} \boldsymbol{\mu}^{*(t)} + (1-w) \cdot \mathbf{F}^{*(0)},
  \vspace{-2mm}
\label{eq:birw2}
\end{align}
where $\mathbf{F}^{*(0)}$ and $\boldsymbol{\mu}^{*(0)}$ are the video snippet features $\mathbf{F}$ and the representative snippets $\boldsymbol{\mu}^{a}$ or $\boldsymbol{\mu}^{e}$, respectively. As shown in Figure \ref{fig:propagation} (bottom), Equations (\ref{eq:birw1}) and (\ref{eq:birw2}) can be also viewed as a EM process, which fixes the affinity $\mathbf{Z}^{*}$ to alternately update $\mathbf{F}$ and $\boldsymbol{\mu}^{*}$. The representative snippets therefore not only are used for propagating representative knowledge (Equation (\ref{eq:birw2})) but also serve as a bridge to propagate the knowledge between the features of $\mathbf{F}$ (Equation (\ref{eq:birw1})). Due to their representativeness, they can better propagate information between features of the same class.
This process can be conducted for multiple times to fully fuse representative snippets' knowledge. To avoid gradient vanishing or explosion caused by the unrolled computational graph, we use an approximate inference formulation as (the detail is available in the supplementary materials)
\begin{equation}\label{eq:approximate}
  \vspace{-1mm}
\resizebox{0.9\columnwidth}{!}
{
  $\bm{{\rm F}}^{*} = (1-w) (\mathbf{I} - w^2 \mathbf{Z}^{*} \operatorname{Norm}_1(\mathbf{Z}^{*})^{\top})^{-1}(w \mathbf{Z}^{*}\boldsymbol{\mu}^{*} + \bm{{\rm F}} ).$
}
\vspace{-1mm}
\end{equation}
%We can obtained that when $\mathbf{F}=\boldsymbol{\mu}^{*}$, \ie propagation is performed between features of $\mathbf{F}$, Equation (\ref{eq:approximate}) falls back to the approximate inference \cite{aldous2002reversible,bertasius2017convolutional,shen2018deep} of the random walk with infinite iterations.
We can also follow Equation (\ref{eq:approximate}) to obtain further refined representative snippets to store in the memory bank. Nevertheless, we find that this way achieves comparable performance with using the original representative snippets. The reason may be that $\boldsymbol{\mu}^{a}$ has been stable enough after several EM iterations. Besides, there is another feasible way to update the video features $\bm{{\rm F}}$ by fixing the representative snippets $\boldsymbol{\mu}^{*}$ and alternately updating $\bm{{\rm F}}$ and $\bm{{\rm Z}}^{*}$. Nevertheless, its performance is comparable with Equations (\ref{eq:birw1}) and (\ref{eq:birw2}) but inferior to Equation (\ref{eq:approximate}).
Note that, rather than concatenating $\boldsymbol{\mu}^{a}$ and $\boldsymbol{\mu}^{e}$ to propagate representative snippets, we use Equation (\ref{eq:approximate}) to propagate the knowledge of $\boldsymbol{\mu}^{a}$ and $\boldsymbol{\mu}^{e}$, respectively. This design is to prevent $\boldsymbol{\mu}^{a}$, which is extracted from the same video of $\mathbf{F}$, being dominated in the propagation. Consequently, after representative snippet propagation, we obtain the updated intra-video features $\mathbf{F}^a$ and the updated inter-video features $\mathbf{F}^e$, respectively.

\subsection{Training Objectives}
Given the original video snippet features $\bm{{\rm F}}$ and the updated features $\bm{{\rm F}}^{a}$, $\bm{{\rm F}}^{e}$, we feed them into three parallel classification heads with sharing parameters to output their TCAMs $\bm{{\rm T}}$, $\bm{{\rm T}}^a$ and $\bm{{\rm T}}^e$, respectively. The TCAMs $\bm{{\rm T}}^a$ and $\bm{{\rm T}}^e$ are weightedly summed to obtain the TCAMs $\bm{{\rm T}}^f$ that contains both intra- and inter-video representative snippet knowledge. We take $\bm{{\rm T}}^f$ as the online pseudo labels for supervising the TCAMs $\bm{{\rm T}}$ as
\begin{equation}\label{eq:soft_supervision}
  \mathcal{L}_{kd} = - \frac{1}{l} \sum\nolimits_{i=1}^l \bm{{\rm T}}^f(i) \mathrm{log}(\bm{{\rm T}}(i)),
\end{equation}
where $t$ is the number of snippets. The total loss is the summation of the loss $\mathcal{L}_{kd}$, the video classification loss $\mathcal{L}_{cls}$ of the three classification heads and the attention normalization loss $\mathcal{L}_{att}$ \cite{zhai2020two} that is only applied to the main branch.
\begin{equation}\label{eq:soft_supervision}
  \mathcal{L} = \mathcal{L}_{cls} + \alpha \mathcal{L}_{kd} + \beta \mathcal{L}_{att},
\end{equation}
where $\alpha$ and $\beta$ are balancing hyper-parameters.

\section{Experiments} \label{sec:experiments}
\subsection{Datasets}

\vspace{-1mm}
\paragraph{THUMOS14. \cite{THUMOS14}}
We use the subset from THUMOS14 that offers frame-wise annotations for 20 classes. We train the model on 200 untrimmed videos in its validation set and evaluate it on 212 untrimmed videos from the test set.

\vspace{-5mm}
\paragraph{ActivityNet1.3. \cite{caba2015activitynet}}
This dataset covers 200 daily activities and provides 10,024 videos for training, 4,926 for validation and 5,044 for testing. We use the training set to train our model and the validation set to evaluate our model.

\subsection{Implementation Details}
\vspace{-1mm}
% \paragraph{Model details.}
% For each snippet, a 2048-$d$ feature is extracted by the I3D pre-trained on Kinetics-400 \cite{carreira2017quo}.
% The following one $1 \times 1$ convolutional layer outputs 2048-$d$ features. Besides, the number of online representative snippets for each video is 8, while the number of representative snippets for each class in the memory bank is 5.

% \vspace{-5mm}
\paragraph{Training details.}
Our method is trained with a mini-batch size of 10 and 16 with Adam \cite{kingma2014adam} optimizer for THUMOS14 and ActivityNet1.3, respectively. The hyper-parameters $w$, $\lambda$, $\alpha$ and $\beta$ are 0.5, 5.0, 1.0 and 0.1, respectively. Since the representative snippets are not well learned at the beginning, we only utilize the intra-video representative snippets in the first 100 epochs, and then add the memory bank into the training process.
The training procedure stops at 200 epochs with the learning rate $5 \times 10^{-5}$.

\vspace{-5mm}
\paragraph{Testing details.}
We take the whole sequence of a video as input for testing. When localizing action instances, the class activation sequence is upsampled to the original frame rate.
We utilize the main branch and the branch of intra-video representative snippets for testing. Their video prediction scores and TCAMs are fused by weighted sum. During detection, We first reject the category whose class probability is lower than 0.1. Following \cite{lee2020background}, we use a set of thresholds to obtain the predicted action instances.
\begin{table}[!t]
\begin{center}
\caption{Results on ActivityNet1.3 validation set. AVG indicates the average mAP at IoU thresholds 0.5:0.05:0.95.}
\vspace{-3mm}
\label{table:activity1.3}
\begin{tabular}{l|cccc}
\hline
\hline
\multirow{2}{*}{Method} & \multicolumn{4}{c}{mAP @ IoU} \\
\cline{2-5}
& 0.5 & 0.75 & 0.95 & AVG \\
\hline
\hline
%R-C3D \cite{xu2017r} & 26.8 & - & - & 12.7 \\
TAL-Net  \cite{chao2018rethinking} & 38.2 & 18.3 & 1.3 & 20.2\\
BSN \cite{lin2018bsn} & 46.5 & 30.0 & 8.0 & 30.0\\
GTAN \cite{long2019gaussian} & 52.6 & 34.1 & 8.9 & 34.3\\
\hline
%STPN (I3D) \cite{nguyen2018weakly} & 29.3 & 16.9 & 2.6 & -\\
%CMCS (I3D) \cite{liu2019completeness} & 34.0 & 20.9 & 5.7 & 21.2\\
%BM (I3D) \cite{nguyen2019weakly} & 36.4 & 19.2 & 2.9 & -\\
BaS-Net (I3D) \cite{lee2020background} & 34.5 & 22.5 & 4.9 & 22.2\\
A2CL-PT (I3D) \cite{min2020adversarial} & 36.8 & 22.0 & 5.2 & 22.5\\
ACM-BANet (I3D) \cite{moniruzzaman2020action} & 37.6 & \textbf{24.7} & \textbf{6.5} & 24.4\\
TSCN (I3D) \cite{zhai2020two} & 35.3 & 21.4 & 5.3 & 21.7\\
WUM (I3D) \cite{lee2021weakly} & 37.0 & 23.9 & 5.7 & 23.7\\
%LES (I3D) \cite{liu2021weakly} & 35.1 & 23.7 & 5.6 & 23.2 \\
TS-PCA (I3D) \cite{liu2021blessings} & 37.4 & 23.5 & 5.9 & 23.7 \\
UGCT (I3D) \cite{yang2021uncertainty} & 39.1 & 22.4 & 5.8 & 23.8 \\
AUMN (I3D) \cite{luo2021action} & 38.3 & 23.5 & 5.2 & 23.5\\
FAC-Net (I3D) \cite{huang2021foreground} & 37.6 & 24.2 & 6.0 & 24.0 \\
Ours (I3D) & \textbf{40.6} & 24.6 & 5.9 & \textbf{25.0} \\
\hline
\hline
\end{tabular}
\vspace{-8mm}
\end{center}
\end{table}
\begin{table*}[!t]
\begin{center}
\caption{Comparisons of detection performance on THUMOS14. UNT and I3D represent UntrimmedNet features and I3D features, respectively. $\dag$ means that the method utilizes additional weak supervisions, \eg, action frequency.}
\label{table:THUMOS14}
\vspace{-3mm}
\resizebox{2\columnwidth}{!}{
\begin{tabular}{c|l|c|ccccccc|ccc}
\hline
\hline
\multirow{2}{*}{Supervision} & \multirow{2}{*}{Method} & \multirow{2}{*}{Feature} & \multicolumn{7}{c|}{mAP @ IoU (\%)} & \multirow{2}{*}{\makecell{AVG\\(0.1:0.5)}} & \multirow{2}{*}{\makecell{AVG\\(0.3:0.7)}} & \multirow{2}{*}{\makecell{AVG\\(0.1:0.7)}} \\
\cline{4-10}
&  &  & 0.1 & 0.2 & 0.3 & 0.4 & 0.5 & 0.6 & 0.7 &  &  & \\
\hline
\hline
\multirow{3}{*}{Full}
%& S-CNN \cite{shou2016temporal}, CVPR2016 & - & 47.7 & 43.5 & 36.4 & 28.7 & 19.0 & 10.3 & 5.3 & 35.0 & 19.9 & 27.3 \\
%& R-C3D \cite{xu2017r}, ICCV2017 & - & 54.5 & 51.5 & 44.8 & 35.6 & 28.9 & - & - & 43.1 & - & - \\
& SSN \cite{zhao2017temporal} (ICCV'17) & - & 60.3 & 56.2 & 50.6 & 40.8 & 29.1 & - & - & 49.6 & - & -\\
%& 2018 & TAL-Net \cite{chao2018rethinking} & 59.8 & 57.1 & 53.2 & 48.5 & 42.8 & 33.8 & 20.8 & 45.1 \\
& BSN \cite{lin2018bsn}  (ECCV'18) & - & - & - & 53.5 & 45.0 & 36.9 & 28.4 & 20.0 & - & 36.8 & - \\
%& BMN \cite{lin2019bmn}, ICCV2019 & - & - & - & 56.0 & 47.4 & 38.8 & 29.7 & 20.5 & - & 38.5 & - \\
& GTAN \cite{long2019gaussian} (CVPR'19) & - & 69.1 & 63.7 & 57.8 & 47.2 & 38.8 & - & - & 55.3 & - & -\\
\hline
\hline
\multirow{2}{*}{Weak $\dag$}
& STAR \cite{xu2019segregated}  (AAAI'19) & I3D & 68.8 & 60.0 & 48.7 & 34.7 & 23.0 & - & - & 47.0 & - & -\\
& 3C-Net \cite{narayan20193c}  (ICCV'19) & I3D & 59.1 & 53.5 & 44.2 & 34.1 & 26.6 & - & 8.1 & 43.5 & - & -\\
\hline
\hline
\multirow{20}{*}{Weak}
%& UntrimmedNet \cite{wang2017untrimmednets}, CVPR2017 & - & 44.4 & 37.7 & 28.2 & 21.1 & 13.7 & - & - & 29.0 & - & - \\
%& Hide-and-Seek \cite{singh2017hide}, ICCV2017 & - & 36.4 & 27.8 & 19.5 & 12.7 & 6.8 & - & - & 20.6 & - & - \\
%& Zhong \emph{et al}. \cite{zhong2018step}, MM2018 & - & 45.8 & 39.0 & 31.1 & 22.5 & 15.9 & - & - & 30.9 & - & - \\
%& AutoLoc \cite{shou2018autoloc}, ECCV2018 & UNT & - & - & 35.8 & 29.0 & 21.2 & 13.4 & 5.8 & - & 21.0 & - \\
& CleanNet \cite{liu2019weakly} (ICCV'19) & UNT & - & - & 37.0 & 30.9 & 23.9 & 13.9 & 7.1 & - & 22.6 & - \\
%& STPN \cite{nguyen2018weakly}, CVPR2018 & I3D & 52.0 & 44.7 & 35.5 & 25.8 & 16.9 & 9.9 & 4.3 & 35.0 & 18.5 & 27.0 \\
%& W-TALC \cite{paul2018w}, ECCV2018 & I3D & 55.2 & 49.6 & 40.1 & 31.1 & 22.8 & - & 7.6 & 39.8 & - & - \\
%& MAAN \cite{yuan2019marginalized}, ICLR2019 & I3D & 59.8 & 50.8 & 41.1 & 30.6 & 20.3 & 12.0 & 6.9 & 40.5 & 22.2 & 31.6 \\
%& CMCS \cite{liu2019completeness}, CVPR2019 & I3D & 57.4 & 50.8 & 41.2 & 32.1 & 23.1 & 15.0 & 7.0 & 40.9 & 23.7 & 32.4 \\
%& BM \cite{nguyen2019weakly}, ICCV2019 & I3D & 60.4 & 56.0 & 46.6 & 37.5 & 26.8 & 17.6 & 9.0 & 36.3 \\
%& BaS-Net \cite{lee2020background}, AAAI2020 & I3D & 58.2 & 52.3 & 44.6 & 36.0 & 27.0 & 18.6 & 10.4 & 43.6 & 27.3 & 35.3 \\
& RPN \cite{huang2020relational} (AAAI'20) & I3D & 62.3 & 57.0 & 48.2 & 37.2 & 27.9 & 16.7 & 8.1 & 46.5 & 27.6 & 36.8 \\
%& DGAM  \cite{shi2020weakly}, CVPR2020 & I3D & 60.0 & 54.2 & 46.8 & 38.2 & 28.8 & 19.8 & 11.4 & 37.0 \\
& TSCN \cite{zhai2020two} (ECCV'20) & I3D & 63.4 & 57.6 & 47.8 & 37.7 & 28.7 & 19.4 & 10.2 & 47.0 & 28.8 & 37.8 \\
& EM-MIL \cite{luo2020weakly} (ECCV'20) & I3D & 59.1 & 52.7 & 45.5 & 36.8 & 30.5 & 22.7 & \textbf{16.4} & 45.0 & 30.4 & 37.7 \\
& A2CL-PT \cite{min2020adversarial} (ECCV'20) & I3D & 61.2 & 56.1 & 48.1 & 39.0 & 30.1 & 19.2 & 10.6 & 46.9 & 29.4 & 37.8 \\
%%& ACM-BANet \cite{moniruzzaman2020action}, MM2020 & I3D & 64.6 & 57.7 & 48.9 & 40.9 & 32.3 & 21.9 & 13.5 & 48.9 & 31.5 & 39.9 \\
& HAM-Net \cite{islam2021hybrid} (AAAI'21) & I3D & 65.4 & 59.0 & 50.3 & 41.1 & 31.0 & 20.7 & 11.1 & 49.4 & 30.8 & 39.8 \\
%& ACSNet \cite{liu2021acsnet} (AAAI'21) & I3D & - & - & 51.4 & 42.7 & 32.4 & 22.0 & 11.7 & - & 32.0 & - \\
%& LES \cite{liu2021weakly}, AAAI2021 & I3D & - & - & 50.8 & 41.7 & 29.6 & 20.1 & 10.7 & - & 30.6 & - \\
& WUM  \cite{lee2021weakly} (AAAI'21) & I3D & 67.5 & 61.2 & 52.3 & 43.4 & 33.7 & 22.9 & 12.1 & 51.6 & 32.9 & 41.9 \\
& AUMN \cite{luo2021action} (CVPR'21) & I3D & 66.2 & 61.9 & 54.9 & 44.4 & 33.3 & 20.5 & 9.0 & 52.1 & 32.4 & 41.5 \\
& CoLA \cite{zhang2021cola} (CVPR'21) & I3D & 66.2 & 59.5 & 51.5 & 41.9 & 32.2 & 22.0 & 13.1 & 50.3 & 32.1 & 40.9 \\
& TS-PCA \cite{liu2021blessings} (CVPR'21) & I3D & 67.6 & 61.1 & 53.4 & 43.4 & 34.3 & 24.7 & 13.7 & 52.0 & 33.9 & 42.6 \\
& UGCT \cite{yang2021uncertainty} (CVPR'21) & I3D & 69.2 & 62.9 & 55.5 & 46.5 & 35.9 & 23.8 & 11.4 & 54.0 & 34.6 & 43.6 \\
& ASL \cite{ma2021weakly} (CVPR'21) & I3D & 67.0 & - & 51.8 & - & 31.1 & - & 11.4 & - & - & - \\
% & CSCL \cite{ji2021weakly} (MM'21) & I3D & 68.0 & 61.8 & 52.7 & 43.3 & 33.4 & 21.8 & 12.3 & 51.8 & 32.7 & 41.9 \\
& CO$_2$-Net \cite{hong2021cross} (MM'21) & I3D & 70.1 & 63.6 & 54.5 & 45.7 & \textbf{38.3} & \textbf{26.4} & 13.4 & 54.4 & 35.6 & 44.6 \\
& D2-Net \cite{narayan2021d2} (ICCV'21) & I3D & 65.7 & 60.2 & 52.3 & 43.4 & 36.0 & - & - & 51.5 & - & - \\
& FAC-Net \cite{huang2021foreground} (ICCV'21) & I3D & 67.6 & 62.1 & 52.6 & 44.3 & 33.4 & 22.5 & 12.7 & 52.0 & 33.1 &  42.2 \\
\cline{2-13}
& \textbf{Ours} & I3D & \textbf{71.3} & \textbf{65.3} & \textbf{55.8} & \textbf{47.5} & 38.2 & 25.4 & 12.5 & \textbf{55.6} & \textbf{35.9} & \textbf{45.1} \\
\hline
\hline
\end{tabular}}
\vspace{-7mm}
\end{center}
\end{table*}

\subsection{Comparison with State-of-the-art Methods}

We compare our method with state-of-the-art weakly supervised methods and several fully supervised methods. The results are shown in Tables \ref{table:activity1.3} and \ref{table:THUMOS14}. On the THUMOS14 dataset, our method evidently outperforms the previous weak-supervised approaches in terms of almost all metrics. On the important criterion: average mAP (0.1:0.5), we surpass the state-of-the-art method \cite{hong2021cross} by 1.2\%.
We also notice that our method achieves better results at IoU 0.1 to 0.5 compared with some recent fully-supervised methods. Even if some methods (\emph{i.e.}, Weak \(\dagger\) in Table \ref{table:THUMOS14}) utilize additional weak supervisions, such as action frequency, our method still outperforms these methods, indicating the effectiveness of our method.
On the larger ActivityNet1.3, our method still outperforms all existing weakly supervised methods by 0.6\% on average mAP.

\subsection{Ablation Study}

We conduct a series of ablation studies on THUMOS14. Unless explicitly stated, we do not utilize the memory bank.
\begin{table}[!t]
\begin{center}
\caption{Evaluation of the necessity and effectiveness of the representative snippet generation. \emph{Baseline} comprises one classification head and takes the video snippet features $\mathbf{F}$ as input.}
\vspace{-3mm}
\label{table:key_frame_generation}
\resizebox{1\columnwidth}{!}{
\begin{tabular}{l|ccccc}
\hline
\hline
\multirow{2}{*}{Method} & \multicolumn{4}{c}{mAP @ IoU} \\
\cline{2-5}
& 0.3 & 0.5 & 0.7 & AVG \\
\hline
\hline
Baseline & 45.2 & 29.9 & 10.2 & 36.8 \\
\hline
+ representative snippets & 48.7 & 33.1 & 11.5 & 40.1 \\
+ pseudo label supervision & 54.5 & 37.3 & 12.5 & 44.2 \\
+ memory bank & 55.8 & 38.2 & 12.5 & 45.1 \\
replace $\boldsymbol{\mu}^{a}$ with $\mathbf{F}$ & 52.4 & 35.5 & 11.7 & 42.4 \\
\hline
\hline
\end{tabular}}
\vspace{-7mm}
\end{center}
\end{table}

\vspace{-9mm}
\paragraph{Representative snippet summarization.}
To validate the effectiveness of the representative snippets, we need to consider: (1) Is generating the representative snippets necessary? (2) If yes, what is an effective way to generate the representative snippets? To answer the two questions, we conduct several  experiments in Tables \ref{table:key_frame_generation} and \ref{table:key_frame_ways}.

We first define a baseline model, which is the same as our main branch that comprises one classification head and also uses the attention normalization loss. As we can see, when we introduce the representative snippets and add the second branch that takes the refined features $\mathbf{{\rm F}}^a$ as input, even if without the supervision between the two branches, the performance is significantly improved over the baseline model. This phenomenon demonstrates the importance of the representative snippets, which makes the model to focus on the representative snippets and thus improves the recognition ability for the whole action class. When we further utilize the TCAMs of the updated features as the online pseudo labels, our method achieves absolute gains of 4.1\% and 7.4\% in terms of average mAP, over the method with only representative snippets and the baseline model.
To further validate the necessity of the representative snippets, we replace representative snippets with the original video features $\mathbf{F}$, and thus the BiRW module becomes a vanilla random walk module. It is noteworthy that the \emph{replace $\boldsymbol{\mu}^{a}$ with $\mathbf{F}$} in Table \ref{table:key_frame_generation} also obtains significant gain over the baseline model, but 1.8\% lower on average mAP than our method using the representative snippets under the same setting.
Besides, our method contains both the intra- and inter-video representative snippets achieves the best performance of 45.1\% on average mAP, indicating the importance of propagating the representative snippets across videos.
\begin{table}[!t]
\begin{center}
\caption{Evaluation of different approaches of generating representative snippets. EM-Att denote the EM attention.}
\vspace{-3mm}
\label{table:key_frame_ways}
\begin{tabular}{l|ccccc}
\hline
\hline
\multirow{2}{*}{Method} & \multicolumn{4}{c}{mAP @ IoU} \\
\cline{2-5}
& 0.3 & 0.5 & 0.7 & AVG \\
\hline
\hline
Discriminative snippets & 45.0 & 25.0 & 6.8 & 34.8 \\
$k$-Means & 51.7 & 34.1 & 11.7 & 41.5 \\
Agglomerative Clustering & 50.0 & 32.3 & 11.2 & 40.1 \\
Spectral Clustering & 50.5 & 31.6 & 11.1 & 40.4 \\
EM-Att w/o BP & 53.1 & 35.2 & 12.1 & 42.8 \\
EM-Att & 54.5 & 37.3 & 12.5 & 44.2 \\
\hline
\hline
\end{tabular}
\vspace{-8mm}
\end{center}
\end{table}

To answer the second questions, we also explore several strategies for representative snippet generation.
As discussed earlier, we first verify the way of selecting discriminative snippets. It is unsurprised that this way deteriorates the performance of the baseline model, because it makes the model only focus on those discriminative snippets and thus weakens the detection ability. In contrast, some traditional clustering methods, such as $k$-means, agglomerative clustering and spectral clustering achieve evident gains over the baseline model. We also evaluate the DBSCAN \cite{ester1996density}. However, as training progresses, it is hard to update proper hyper-parameters for DBSCAN, leading to very low performance.
It is noteworthy that the \emph{EM-Att w/o BP} achieves a high performance of 42.8\% in terms of average mAP, which is much better than other clustering methods. It may attribute to its orthogonal initialization, which enables the learned representative snippets to concentrate on different video parts. Besides, the performance is further improved by 1.4\% when we allow back propagation. More experimental results are provided in the supplementary materials.

%In the supplementary materials, we provide more experimental results of the representative snippets.
\begin{table}[!t]
\begin{center}
\caption{Evaluation of the representative snippet propagation.}
\vspace{-3mm}
\label{table:key_frame_propagation}
\resizebox{1.0\columnwidth}{!}{
\begin{tabular}{l|ccccc}
\hline
\hline
\multirow{2}{*}{Method} & \multicolumn{4}{c}{mAP @ IoU} \\
\cline{2-5}
& 0.3 & 0.5 & 0.7 & AVG \\
\hline
\hline
Random walk $\bm{{\rm Z}}^{a}\boldsymbol{\mu}^{a}$ & 45.0 & 25.0 & 6.8 & 34.8 \\
Equation (\ref{eq:approximate}) & 54.5 & 37.3 & 12.5 & 44.2 \\
\hline
Propagate scores & 49.5 & 28.8 & 8.2 & 38.5 \\
Propagate features w/o $\mathcal{L}_{cls}$ & 48.2 & 28.4 & 8.3 & 37.7 \\
\hline
\hline
\end{tabular}}
\vspace{-8mm}
\end{center}
\end{table}
\begin{figure}[!t]
  \centering
  \includegraphics[width=0.9\columnwidth]{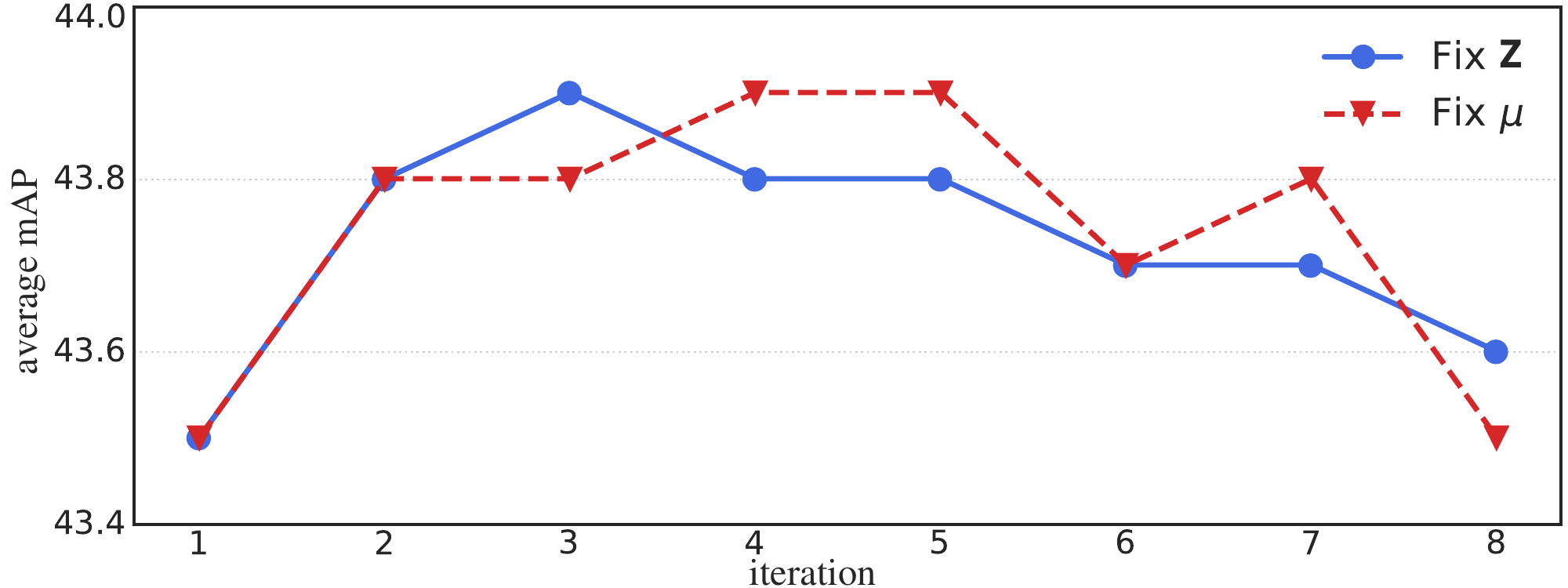}
  \caption{The detection results of two implementations of Equations (\ref{eq:birw1}) and (\ref{eq:birw2}) with different iteration numbers.}\label{fig:em_iter_num}
\vspace{-2mm}
\end{figure}

\vspace{-6mm}
\paragraph{Representative snippet propagation.}
Several comparative experiments are conducted to evaluate the effectiveness of representative snippet propagation, with results posted in Table \ref{table:key_frame_propagation}. We first evaluate some variants of the representative snippet propagation. We notice that the random walk $\bm{{\rm Z}}^{a}\boldsymbol{\mu}^{a}$ deteriorates the performance. In contrast, as shown in Figure \ref{fig:em_iter_num}, even if we weightedly sum $\bm{{\rm Z}}^{a}\boldsymbol{\mu}^{a}$ with the original features (\ie, iteration 1), the performance would be significantly improved, indicating that $\bm{{\rm Z}}^{a}\boldsymbol{\mu}^{a}$ deviates too much from the original video features $\mathbf{F}$. Moreover, the performance improves as the number of iterations increases. However, when the number of iterations is too large, the performance slightly drops,which might be caused by gradient vanishing or explosion even if there is residual mechanism (weighted summation with the video features). Despite their promising performance, our proposed approximation inference obtain the optimal result.

We further validate two more variants of propagation. The first variant, \emph{propagate scores}, follows a knowledge ensembling way to propagate prediction scores of the representative snippets rather than their features. Therefore, only the main branch has the video-level classification losses. The second variant, \emph{propagate features w/o $\mathcal{L}_{cls}$}, also propagates the features of representative snippets but removes the video-level classification losses. Both variants have gains over the baseline model but are much inferior to our final solution. Therefore, the two additional branches not only provide online pseudo labels, but also enforce the model to focus on the representative snippets, thus improving the model's ability to localize action instances.
\begin{table}[!t]
\begin{center}
\caption{The detection results of applying our method to existing methods. \emph{Embedding} means we add a learnable network after the backbone network to learn the video features. The results of the original methods are reproduced.}
\vspace{-3mm}
\label{table:improve_existing_methods}
\resizebox{1\columnwidth}{!}{
\begin{tabular}{l|ccccl}
\hline
\hline
\multirow{2}{*}{Method} & \multicolumn{4}{c}{mAP @ IoU} \\
\cline{2-5}
& 0.3 & 0.5 & 0.7 & AVG \\
\hline
\hline
STPN \cite{nguyen2018weakly} + embedding & 38.4 & 19.1 & 4.7 & 28.9 \\
STPN + embedding + Ours & 46.5 & 21.8 & 5.8 & 33.7 $_{\uparrow 4.8}$ \\
\hline
BM \cite{nguyen2019weakly} & 45.2 & 26.2 & 8.7 & 35.3 \\
BM + Ours & 47.4 & 29.3 & 9.2 & 37.7 $_{\uparrow 2.4}$ \\
\hline
WUM \cite{lee2021weakly} & 51.0 & 32.8 & 10.9 & 40.4 \\
WUM + Ours & 53.5 & 34.0 & 12.7 & 42.4 $_{\uparrow 2.0}$ \\
\hline
FAC-Net \cite{huang2021foreground} & 53.2 & 34.4 & 13.7 & 42.9 \\
FAC-Net + Ours & 55.8 & 38.4 & 13.3 & 45.2 $_{\uparrow 2.3}$ \\
\hline
\hline
\end{tabular}}
\vspace{-8mm}
\end{center}
\end{table}

\vspace{-5mm}
\paragraph{Integrating our framework to existing methods.}
In Table \ref{table:improve_existing_methods}, we replace our classification heads with some existing methods. In these experiments, we use the default settings of these methods to ensure fair comparisons. Besides, if there are additional losses in these methods, we only impose them on the main branch. As we can see, either for some early methods or state-of-the-art methods, our method can consistently improve their performances.
\begin{figure}[!t]
  \centering
  \includegraphics[width=1.0\columnwidth]{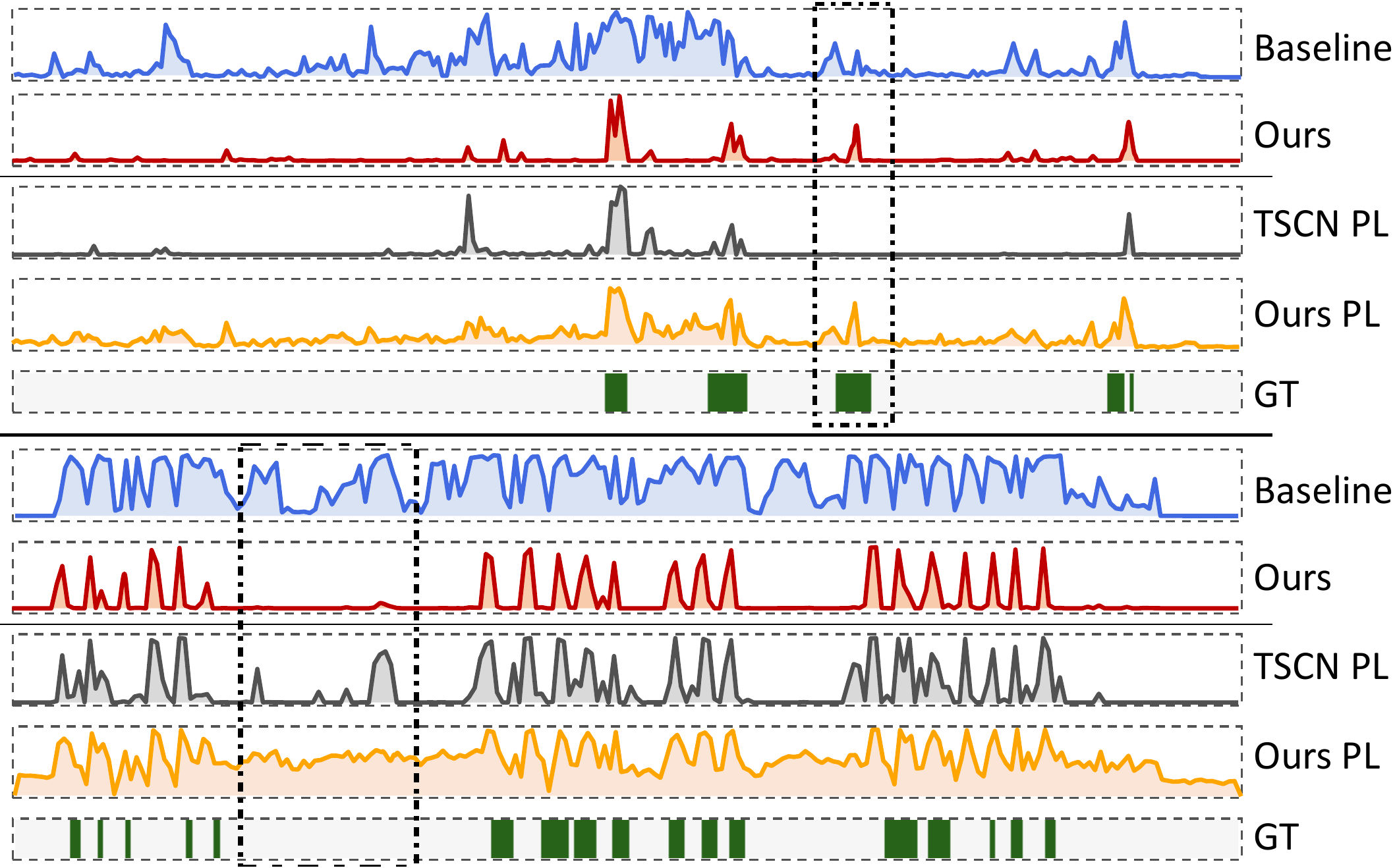}
\caption{We show the qualitative results of two examples of \emph{shotput} and \emph{diving} on THUMOS14 \cite{THUMOS14}. PL denotes the pseudo label.}
\label{fig:detection_visualization}
\vspace{-4mm}
\end{figure}

\subsection{Qualitative results}
We visualize some examples of detected action instances and generated pseudo labels in Figure \ref{fig:detection_visualization}. In the first example of \emph{shotput}, compared with TSCN \cite{zhai2020two}, our method generates more accurate pseudo labels to successfully detect the third action instance (black rectangle). In the second example of \emph{diving}, our method leverages the knowledge of the representative snippets of background to effectively suppress the responses of the background.

\vspace{-1mm}
\section{Conclusion and Limitation}
\vspace{-1mm}
We propose a representative snippet summarization and propagation framework. Our method aims to generate better pseudo labels by propagating the knowledge of representative snippets.
We summarize the representative snippets in each video and maintain a memory table to store representative snippets. For each video, the intra- and inter-video representative snippets are propagated to update the video features. The temporal class activation maps of the updated features serve as the online pseudo labels to rectify the predictions of the main branch.
Our method achieves the state-of-the-art performance on two popular datasets and can consistently improve the performance of existing methods.

Due to the matrix inversion during propagation, the training time of one epoch in our method is almost twice as long as the baseline model. We will resort to some efficient matrix inversion strategies to solve this issue in the future.

\section*{Acknowledgement}
This work is supported in part by Centre for Perceptual and Interactive Intelligence Limited, in part by the General Research Fund through the Research Grants Council of Hong Kong under Grants (Nos. 14204021, 14207319), in part by CUHK Strategic Fund.

%%%%%%%%% REFERENCES
{\small
\bibliographystyle{ieee_fullname}
\bibliography{egbib}
}

\end{document}

%%%%%%%%% TITLE - PLEASE UPDATE
\title{Supplementary Material}

\author{Linjiang Huang\textsuperscript{\rm 1,2} \quad
        Liang Wang\textsuperscript{\rm 3} \quad
        Hongsheng Li\textsuperscript{\rm 1,2} \thanks{Corresponding author.}\\
\textsuperscript{\rm 1}Multimedia Laboratory, The Chinese University of Hong Kong \\
\textsuperscript{\rm 2}Centre for Perceptual and Interactive Intelligence, Hong Kong \\
\textsuperscript{\rm 3}Institute of Automation, Chinese Academy of Sciences \\
{\tt\small ljhuang524@gmail.com, wangliang@nlpr.ia.ac.cn, hsli@ee.cuhk.edu.hk}
}

\maketitle

%%%%%%%%% ABSTRACT
In this supplementary material, more details about our methods are presented first. Then we report additional experimental results to further validate our network design. At last, we show more qualitative detection results, some failure cases and feature visualizations.

%%%%%%%%% BODY TEXT

\section{Architecture of the Classification Head}
The classification head is mainly used to generate TCAMs, it can be any existing WSTAL methods. To generate high quality TCAMs and improve the lower bound of our method, we use the recent method FAC-Net
\cite{huang2021foreground} as the classification head for its simple pipeline and strong performance.

\begin{figure}
  \centering
  \includegraphics[width=1\columnwidth]{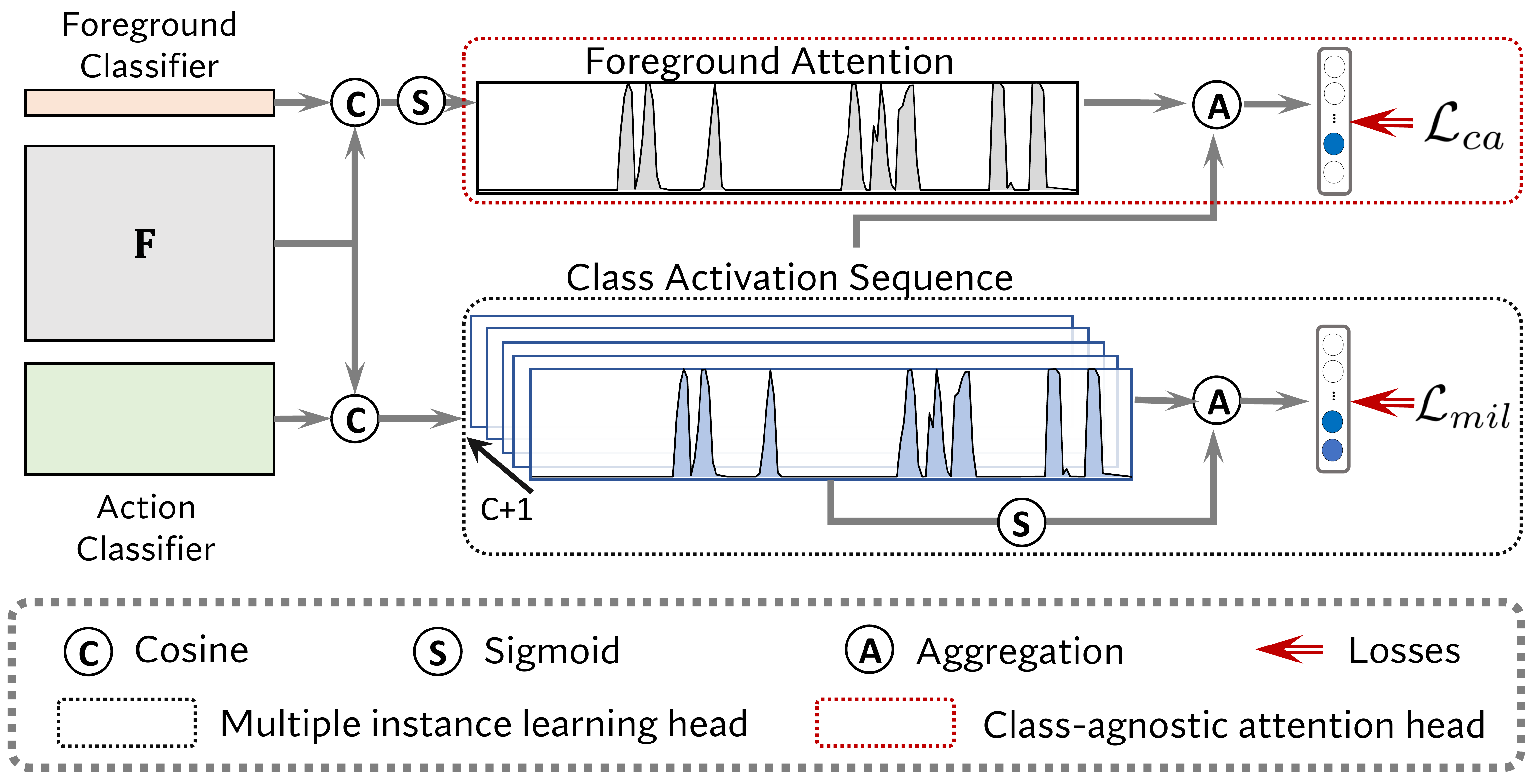}
  \caption{The overview of the classification head used in our method, including a class-agnostic attention head and a multiple instance learning head in FAC-Net \cite{huang2021foreground}.}\label{fig:classification_head}
\vspace{-5mm}
\end{figure}

We simply revisit the FAC-Net in this section. As the most weakly supervised methods, FAC-Net is based on the fixed-weight I3D backbone network, on which a small learnable network (\eg, $1 \times 1$ convolutional network) is appended to learn the video snippet features $\mathbf{F} \in \mathbb{R}^{l \times d}$.
Besides, FAC-Net contains a foreground classifier \(\bm{{\rm W}}_f\) and an action classifier \(\bm{{\rm W}}_a\). Given the video snippet features $\mathbf{F}$, there are three classification heads are appended on.
The first classification head is a class-agnostic attention (CA) head. It first calculates the cosine similarity between \(\bm{{\rm W}}_f\) and \(\bm{{\rm F}}\) to get the foreground attention weights \(\bm{{\rm \lambda}}_f \in  \mathbb{R}^{l}\). The foreground attention weights are used to aggregate snippet features \(\bm{{\rm F}}\) into a video level feature \(\bm{{\rm \bar{F}}}\), which is used to calculate the video-level prediction with \(\bm{{\rm W}}_a\).
The second classification head is a multiple instance learning (MIL) head. It calculates the cosine similarity between \(\bm{{\rm W}}_a\) and \(\bm{{\rm F}}\) to obtain the snippet-wise class logits \(\bm{{\rm S}}  \in  \mathbb{R}^{l \times (c+1)}\). The TCAMs \(\bm{{\rm T}}\) and the class-wise attention scores \(\bm{{\rm \lambda}}_w\) are obtained by applying softmax to \(\bm{{\rm S}}\) along category dimension and temporal dimension, respectively. The video-level prediction of the MIL head is obtained by aggregating \(\bm{{\rm S}}\) with \(\bm{{\rm \lambda}}_w\). There is another classification head, namely class-wise foreground classification head. However, as shown in Figure \ref{fig:classification_head}, we do not use this head in our method, since the former two heads already enable our method to achieve a high baseline performance. Therefore, in our method, for each classification head, there are only two video-level classification losses, which corresponds to the CA branch and the MIL branch, respectively. We weightedly combine the two video-level classification losses as $\mathcal{L}_{cls} = \mathcal{L}_{ca} + \gamma \mathcal{L}_{mil}$, where $\gamma$ denotes the balancing hyper-parameter, which is set as 0.2 in our method.

There are some other modifications upon the FAC-Net. First, we use the sigmoid rather than the softmax function to obtain normalized foreground scores. This setting enables our method to use the attention normalization term \cite{zhai2020two} to obtain highly confident representative snippets. Second, we do not use the hybrid attention strategy, which is designed to alleviate the discrepancy between classification and detection.

\section{Bipartite Random Walk Module}
First, we briefly introduce the random walk operation. Random walks are one of the most widely known and used methods in graph theory \cite{lovasz1993random}, which led to the development of PageRank \cite{page1999pagerank}, personalized PageRank \cite{bahmani2010fast}, DeepWalk \cite{perozzi2014deepwalk}, DeeperGCN \cite{klicpera2018predict}, \etc. Let $G$ denotes an undirected graph. There is a transition matrix $\mathbf{A}$ of the graph $G$. Given the feature $\mathbf{F}$ of graph nodes, the random walk operation can be done via a simple matrix multiplication operation $\mathbf{F}^{(1)} = \mathbf{A}\mathbf{F}$.

In \cite{bertasius2017convolutional,shen2018deep}, they use the random walk operation to propagate information between image pixels or probe images for semantic segmentation and person re-identification. They propose to weighted combine $\mathbf{A}\mathbf{F}$ with the input $\mathbf{F}$ to make $\mathbf{F}^{(1)}$ be not deviate too far away from $\mathbf{F}$, which can be formulated as
\begin{equation}
  \mathbf{F}^{(1)} = w\mathbf{A}\mathbf{F} + (1-w) \mathbf{F},
\end{equation}
where $w$ is a balancing hyper-parameter. This process can be conducted multiple times until convergence
\begin{equation}
  \mathbf{F}^{(t)} = w\mathbf{A}\mathbf{F}^{(t-1)} + (1-w) \mathbf{F}.
\end{equation}
As $t \rightarrow \infty$, there is an approximate inference as
\begin{equation}  \label{eq:rw_approximate}
  \mathbf{F}^{(\infty)} = (1-w)(\mathbf{I}-w\mathbf{A})^{-1}\mathbf{F},
\end{equation}
where $\mathbf{I}$ denotes an identity matrix.

As state in our main manuscript, at the $t$-th iteration, the bipartite random walk operation can be formulated as (for simplicity, we omit the upper-script *, which is $a$ or $e$ corresponding to the online representative snippets or the offline representative snippets)
\begin{align}\label{eq:birw1}
  \vspace{-1mm}
  \boldsymbol{\mu}^{(t)} =& w \operatorname{Norm}_1(\mathbf{Z})^{\top} \mathbf{F}^{(t-1)} + (1-w) \boldsymbol{\mu}^{(0)}, \\
  \mathbf{F}^{(t)} =& w \mathbf{Z}\boldsymbol{\mu}^{(t)} + (1-w) \mathbf{F}^{(0)}.
  \vspace{-1mm}
\label{eq:birw2}
\end{align}
Substitute Equation (\ref{eq:birw1}) into Equation (\ref{eq:birw2}), we can obtain
\begin{eqnarray} \label{eq:substitute}
\mathbf{F}^{(t)} &=& w^2 \mathbf{Z} \operatorname{Norm}_1(\mathbf{Z})^{\top} \mathbf{F}^{(t-1)} + (1-w)w  \mathbf{Z} \boldsymbol{\mu}^{(0)} \nonumber  \\
&\;& + (1-w) \mathbf{F}^{(0)}  \nonumber \\
&=& w^2 \mathbf{Z} \operatorname{Norm}_1(\mathbf{Z})^{\top} \mathbf{F}^{(t-1)} \nonumber  \\
&\;& + (1-w) (w\mathbf{Z} \boldsymbol{\mu}^{(0)} + \mathbf{F}^{(0)}).
\end{eqnarray}
In the following equations, we denote $\mathbf{Z} \operatorname{Norm}_1(\mathbf{Z})^{\top}$ as $\mathbf{R}$. Therefore, expanding Equation (\ref{eq:substitute}) leads to
\begin{equation}
  \mathbf{F}^{(t)} = (w^2 \mathbf{R})^{t}\mathbf{F}^{(0)}+(1-w)\sum_{i=0}^{t-1}(w^2\mathbf{R})^{i}(w\mathbf{Z} \boldsymbol{\mu}^{(0)} + \mathbf{F}^{(0)}).
\end{equation}
As $t \rightarrow \infty$, since elements in $\mathbf{R}$ and $w \in [0, 1]$,
\begin{equation}
  \lim_{t \rightarrow \infty}(w^2 \mathbf{R})^{t}\mathbf{F}^{(0)} = \bm{0}.
\end{equation}
For $\sum_{i=0}^{t-1}(w^2\mathbf{R})^{i}$, the matrix series can be expanded as
\begin{equation}
  \lim_{t \rightarrow \infty}\sum_{i=0}^{t-1} (w^2\mathbf{R})^{i} = (\mathbf{I}- w^2\mathbf{R})^{-1}.
\end{equation}
Therefore, Equation (\ref{eq:substitute}) can be formulated as
\begin{equation}\label{eq:approximate}
\resizebox{1.0\columnwidth}{!}
{
  $\bm{{\rm F}}^{(\infty)} = (1-w) (\mathbf{I} - w^2 \mathbf{Z} \operatorname{Norm}_1(\mathbf{Z})^{\top})^{-1}(w \mathbf{Z}\boldsymbol{\mu}^{(0)} + \bm{{\rm F}}^{(0)} ).$
}
\end{equation}
When $\boldsymbol{\mu}^{(0)}=\bm{{\rm F}}^{(0)}$, that is, the propagation is directly conducted between the features of $\mathbf{F}$, there exists $\operatorname{Norm}_1(\mathbf{Z})^{\top}=\mathbf{Z}$, because the bi-directional affinities between $\bm{{\rm F}}^{(0)}$ and $\bm{{\rm F}}^{(0)}$ should be the same. At this time, Equation (\ref{eq:approximate}) can be rewrote as
\begin{equation}\label{eq:approximate2}
\begin{split}
\bm{{\rm F}}^{(\infty)} &= (1-w) (\mathbf{I} - w^2 \mathbf{Z}^2)^{-1}(w \mathbf{Z} \bm{{\rm F}} ^{(0)} + \bm{{\rm F}}^{(0)} ), \\
     &= (1-w)(\mathbf{I} - w \mathbf{Z})^{-1}(\mathbf{I} + w \mathbf{Z})^{-1}(w \mathbf{Z} \bm{{\rm F}} ^{(0)} + \bm{{\rm F}}^{(0)} ), \\
     &= (1-w)(\mathbf{I} - w \mathbf{Z})^{-1}\bm{{\rm F}} ^{(0)},
\end{split}
\end{equation}
which is the same as Equation (\ref{eq:rw_approximate}).

\section{More Implementation Details}
For each snippet, a 2048-$d$ feature is extracted by the I3D pre-trained on Kinetics-400 \cite{carreira2017quo}. The following one $1 \times 1$ convolutional layer outputs 2048-$d$ features. Besides, the number of online representative snippets for each video is 8, while the number of representative snippets for each class in the memory bank is 5.

\section{Additional Ablation Study}
In this section, we provide more ablation studies on the THUMOS14 dataset. Unless explicitly stated, we do not use the memory bank in these methods.
\begin{table}[!t]
\begin{center}
\caption{Evaluation of the attention normalization loss.}
\vspace{-3mm}
\label{table:attention_loss}
\begin{tabular}{l|cc}
\hline
\hline
Method & $\mathcal{L}_{att}$ & AVG \\
\hline
\hline
\multirow{2}{*}{Baseline} & \XSolidBrush & 35.6 \\
& \CheckmarkBold & 36.8 \\
\hline
\multirow{2}{*}{+ representative snippets} & \XSolidBrush & 40.2 \\
& \CheckmarkBold & 40.1 \\
\hline
\multirow{2}{*}{+ pseudo label supervision} & \XSolidBrush & 42.0 \\
& \CheckmarkBold & 44.2 \\
\hline
\multirow{2}{*}{+ memory bank} & \XSolidBrush & 42.5 \\
& \CheckmarkBold & 45.1 \\
\hline
\hline
\end{tabular}
\vspace{-8mm}
\end{center}
\end{table}

\vspace{-5mm}
\paragraph{Attention normalization loss.} In Table \ref{table:attention_loss}, we demonstrate the effectiveness of the attention normalization loss. As we can see, the attention normalization loss plays an important role in our method. It improves the performances of our baseline model as well as the full model. Besides, we can see that the full model without the attention normalization loss drops evidently, indicating that the attention normalization loss is essential to obtain useful representative snippets.
\begin{table}[!t]
\begin{center}
\caption{The detection results of different variants of generating representative snippets.}
\vspace{-3mm}
\label{table:representative snippets_summarization}
\resizebox{1\columnwidth}{!}{
\begin{tabular}{l|ccccl}
\hline
\hline
\multirow{2}{*}{Method} & \multicolumn{4}{c}{mAP @ IoU} \\
\cline{2-5}
& 0.3 & 0.5 & 0.7 & AVG \\
\hline
\hline
Baseline & 45.2 & 29.9 & 10.2 & 36.8 \\
Full model w/o memory bank & 54.5 & 37.3 & 12.5 & 44.2 \\
\hline
Foreground representative snippet & 52.2 & 33.2 & 11.7 & 41.8 \\
background representative snippet & 51.5 & 30.9 & 8.9 & 40.2 \\
\hline
representative snippets of high scores & 49.2 & 30.2 & 8.5 & 39.2 \\
\hline
video-wise memory bank & 56.2 & 38.2 & 12.4 & 45.0 \\
\hline
\hline
\end{tabular}}
\vspace{-7mm}
\end{center}
\end{table}

\vspace{-5mm}
\paragraph{Representative snippet summarization.} Here, we provide more experimental results to evaluate the representative snippet summarization. In our method, we generate the representative snippets regardless of the foreground or background. Since the classification head generates foreground scores, we can also force the model to only generate representative snippets of the foreground or background. Specifically, given the foreground scores $\mathbf{S}_f$ of the main branch, we first modulate the video snippet features $\mathbf{F}$ with $\mathbf{S}_f$ as $\mathbf{F}_f=\mathbf{F}\mathbf{S}_f$, and then use the EM attention to obtain the representative snippets based on the modulated features $\mathbf{F}_f$. We update the foreground features $\mathbf{F}_f$ by the bipartite random walk module, and finally summed up with the background features $\mathbf{F}_b=\mathbf{F}(1-\mathbf{S}_f)$ to obtain the updated features. For the background representative features, we can follow the same pipeline to obtain the updated features.

In Table \ref{table:representative snippets_summarization}, we can see that, generating only foreground or background representative snippets cannot model the whole video, and thus the pseudo labels of background or foreground are still inaccurate. Besides, we also try to simultaneously generate foreground and background representative snippets by selecting the top-$k$ and the bottom-$k$ representative snippets according to their foreground scores. As we can see, this way obtains much inferior performance. Since different videos can be differently complex, manually selecting $k$ foreground representative snippets and $k$ background representative snippets is not enough to capture the variations of the whole video.

\vspace{-5mm}
\paragraph{Memory bank.}
In Table \ref{table:representative snippets_summarization}, we also evaluate a variant of the memory bank. For each video, we use a separated memory bank to store all the representative snippets. During training, we randomly select one video of the same class from the dataset and retrieve its representative snippets as the offline representative snippets $\boldsymbol{\mu}^e$. We update the representative snippets in the memory bank by moving average just as the MoCo \cite{he2020momentum}. As we can see, this way achieves comparable performance with ours, indicating the effectiveness of propagating knowledge between videos. However, it requires much more memory (about $15\times$) to store the representative snippets.

\vspace{-4mm}
\paragraph{Representative snippet propagation.}
Here, we provide more experimental results to evaluate the representative snippet propagation, the experimental results are shown in Table \ref{table:representative snippets_propagation}. We first evaluate the way of propagating the representative snippets, \ie, the feature L1 loss and feature L2 loss in Table \ref{table:representative snippets_propagation}. Specifically, the feature L1 loss denotes that we enforce the original video features $\mathbf{F}$ and updated features $\mathbf{F}^a$ to be similar by L1 loss, \ie, $\mathcal{L}_{l1} = \|\mathbf{F} - \mathbf{F}^a\|_1$. Likewise, the feature L2 loss denotes $\mathcal{L}_{l2} = \|\mathbf{F} - \mathbf{F}^a\|_2^2$. We can see that, both methods deteriorate the performance. This phenomenon may be due to the optimization objective of the two loss functions is not the prediction scores but the features, which cannot directly guarantee consistent scores to obtain desirable detection results.

Moreover, in our method, we adopt the late fusion manner that combines the predictions of features updated with the online and offline representative snippets. Here, we also evaluate the early fusion strategy, which first concatenates the online and offline representative snippets and then propagate their knowledge to update the original video features and generate predictions. As we can see, this method obtains inferior performance, which may be due to the dominance of the online representative snippets in propagation.

\begin{table}[!t]
\begin{center}
\caption{The detection results of different variants of representative snippet propagation.}
\vspace{-3mm}
\label{table:representative snippets_propagation}
\begin{tabular}{l|ccccl}
\hline
\hline
\multirow{2}{*}{Method} & \multicolumn{4}{c}{mAP @ IoU} \\
\cline{2-5}
& 0.3 & 0.5 & 0.7 & AVG \\
\hline
\hline
Feature L1 loss & 52.1 & 32.1 & 10.1 & 41.1 \\
Feature L2 loss & 52.7 & 31.6 & 10.2 & 41.2 \\
\hline
Concat $\boldsymbol{\mu}^a$ and $\boldsymbol{\mu}^e$ & 54.4 & 35.2 & 10.6 & 43.0 \\
\hline
\hline
\end{tabular}
\vspace{-6mm}
\end{center}
\end{table}
\begin{figure}
  \centering
  \includegraphics[width=0.95\columnwidth]{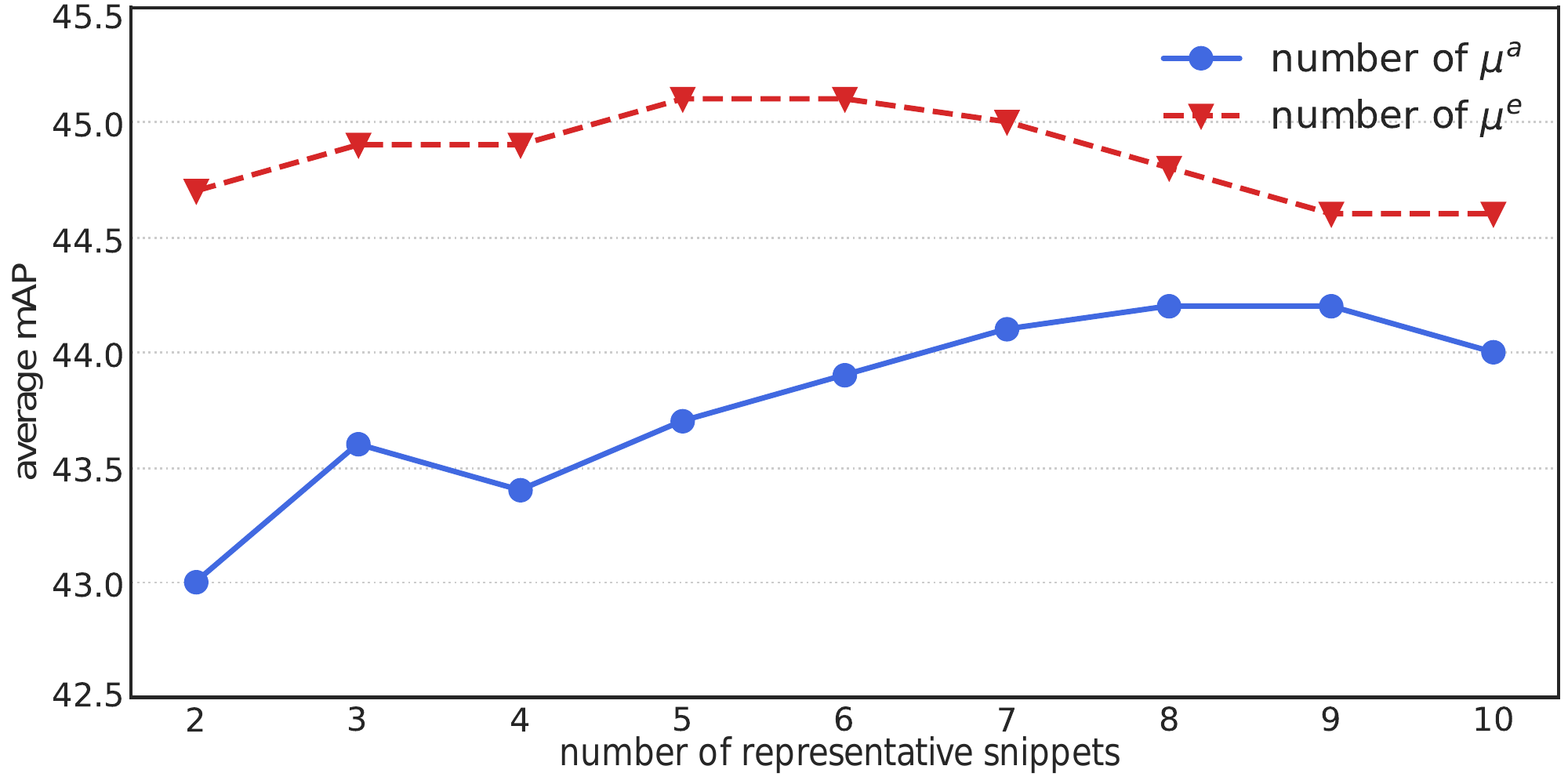}
  \caption{Ablation study on the numbers of representative snippets $\boldsymbol{\mu}^{a}$ and $\boldsymbol{\mu}^{e}$.}\label{fig:num_snippet}
  \vspace{-3mm}
\end{figure}
\begin{figure}
  \centering
  \includegraphics[width=0.95\columnwidth]{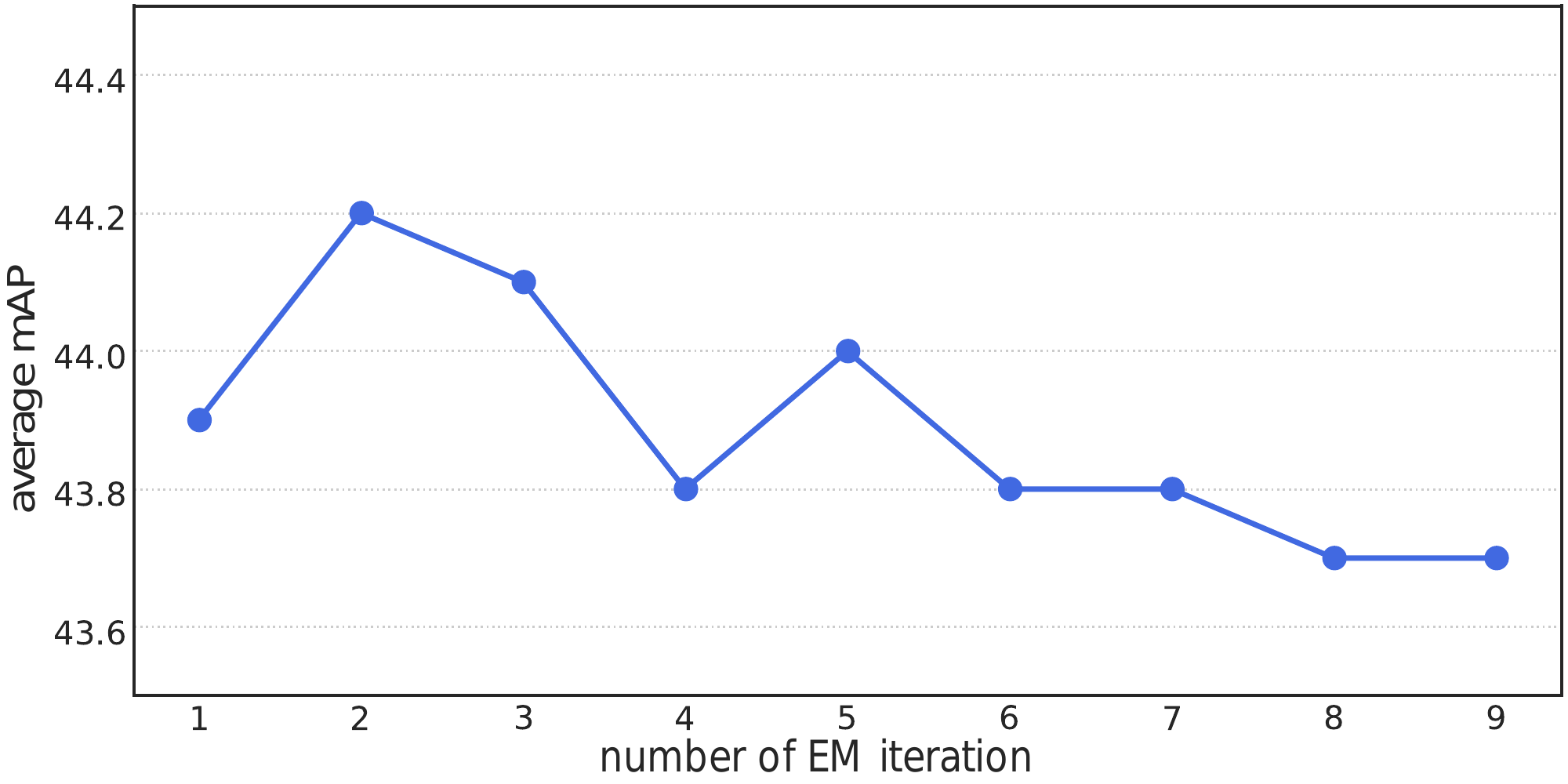}
  \caption{Ablation study on the number of EM iterations.} \label{fig:num_em}
  \vspace{-5mm}
\end{figure}

\begin{table}[!t]
\begin{center}
\caption{Evaluation of three branches of their detection results and MACs. There are also the results of two methods W-TALC \cite{paul2018w} and CO$_2$-Net \cite{hong2021cross} are demonstrated for comparison. All MACs are calculated with the same video having 100 frames and do not contain those of the fixed I3D.}
\vspace{-3mm}
\label{table:branch_analysis}
\resizebox{1\columnwidth}{!}{
\begin{tabular}{l|cccc|c}
\hline
\hline
\multirow{2}{*}{Method} & \multicolumn{4}{c|}{mAP @ IoU} & \multirow{2}{*}{MACs} \\
\cline{2-5}
& 0.3 & 0.5 & 0.7 & AVG \\
\hline
\hline
W-TALC \cite{paul2018w} & 40.1 & 22.8 & 7.6 & - & 0.42G\\
CO$_2$-Net \cite{hong2021cross} & 54.5 & 38.3 & 13.4 & 44.6 & 2.79G \\
\hline
Main branch & 55.5 & 38.2 & 12.5 & 44.7 & 0.44G \\
\hline
+ intra-video branch & 55.8 & 38.2 & 12.5 & 45.1 & 0.48G \\
+ inter-video branch & 55.9 & 38.2 & 12.5 & 45.2 & 0.90G \\
\hline
\hline
\end{tabular}}
\vspace{-7mm}
\end{center}
\end{table}

\vspace{-5mm}
\paragraph{Branch analysis.}
In our method, there are three branches, namely the main branch, the intra-video branch and the inter-video branch. In Table \ref{table:branch_analysis}, we show their performance and MACs. First, we should highlight that inference can be done with only the main branch. Note that, all existing methods (including ours) use a fixed I3D (MACs 5500G). Except I3D's MACs, with only the main branch, our MACs (0.44G) is on par with W-TALC \cite{paul2018w}, but achieves 44.7\% on THUMOS14, which is still higher than the best model CO$_2$-Net \cite{hong2021cross} (44.6\%, MACs 2.79G) by 0.1\% but with much less computational cost.
Besides, snippet summarization, propagation and the intra-video branch incorporate little overhead, the main computation cost is the feature extraction module, which accounts for about 88\% of the whole computation cost. Therefore, after adding the intra-video branch, our model (45.1\%, MACs 0.48G) is still much more efficient than previous methods. Finally, we can see that further adding the inter-video branch can also improve the performance. However, since we do not know the video classes, we should first perform a round of inference to obtain the video prediction and then select the inter-video snippets according to the predicted video classes for the second round inference, which significantly increases the inference time. Therefore, we do not use the inter-video branch during testing in our final solution.

\vspace{-5mm}
\paragraph{Number of representative snippets.} In Figure \ref{fig:num_snippet}, we evaluate the impact of the number of representative snippets on our method. When we evaluate the number of online representative snippets $\boldsymbol{\mu}^a$, we do not adopt the memory bank. When we evaluate the number of offline representative snippets $\boldsymbol{\mu}^e$, we fix the number of online representative snippet as the optimal value. We can see that our method can benefit much from the online representative snippets, even if we only utilize two Gaussians to model the foreground and background in the video, the performance is much higher than the baseline model (average mAP 36.8\%). Our method achieves the optimal performance when the number of online representative snippets is 8. Moreover, our method is insensitive to the number of offline representative snippets, even 2 offline representative snippets can enable our method to achieve a promising performance. Our method achieves the optimal performance when the number of offline representative snippets of each class in the memory table is 5.

\vspace{-5mm}
\paragraph{Number of EM iterations.}
In Figure \ref{fig:num_em}, we evaluate the impact of the number of EM iterations on our method. As we can see, our method obtains promising representative snippets by 2 iterations. When the number of EM iterations increases, the performance slightly drop, which might be caused by gradient vanishing or explosion.
\begin{figure}
  \centering
  \includegraphics[width=1\columnwidth]{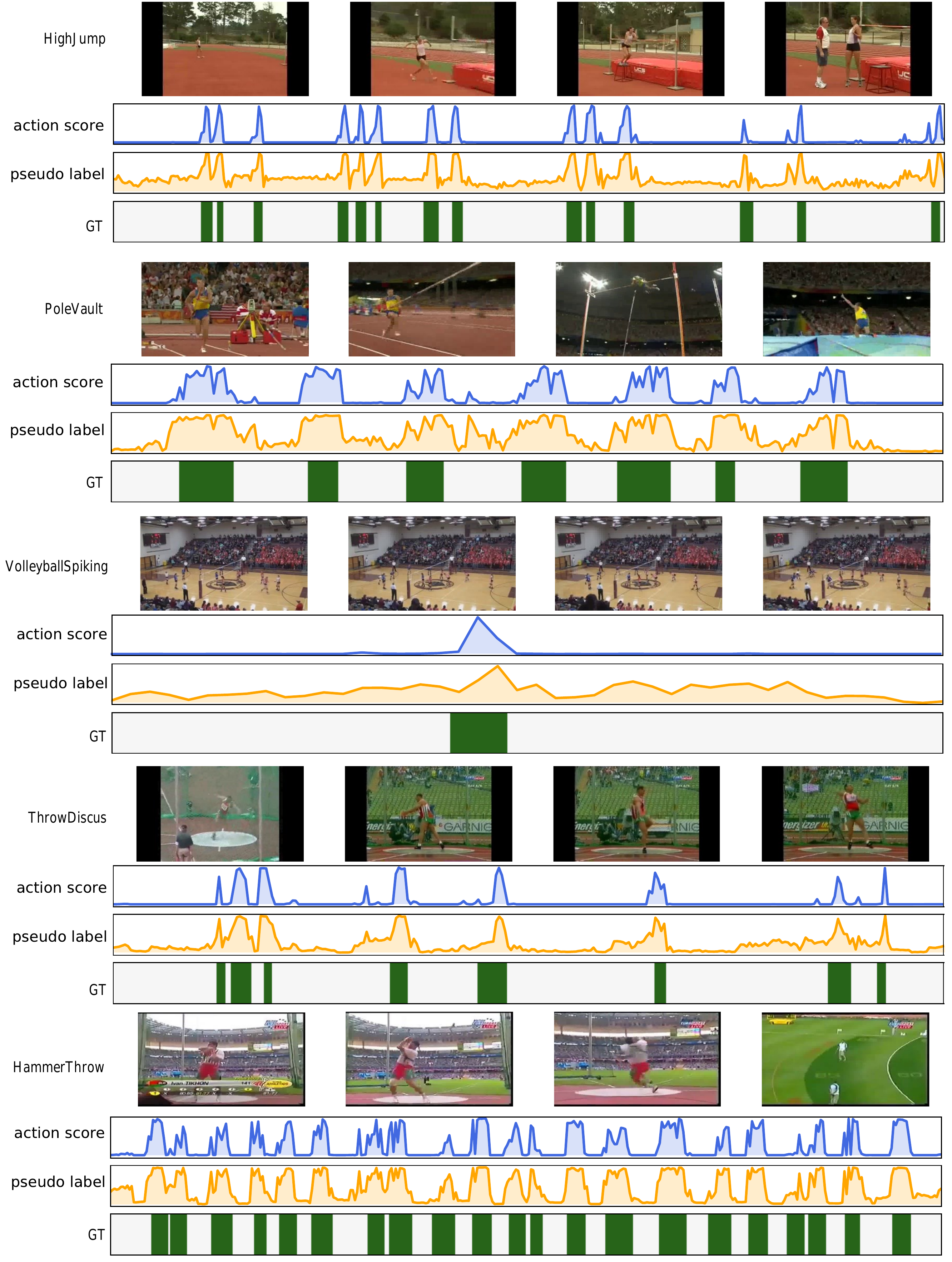}
  \caption{Qualitative results on THUMOS14 \cite{THUMOS14}. We show: 1) action activation scores, 2) pseudo label scores of ground truth action, 3) ground truth.} \label{fig:detection_visualization}
  \vspace{-5mm}
\end{figure}

\subsection{Qualitative Results}

\paragraph{Detection results.}
We visualize some detection examples in Figure \ref{fig:detection_visualization}. We provide some frames of the input videos to show the corresponding actions. As we can see, for videos containing sparse or dense action instances, the localization results of our method are complete and accurate.
\begin{figure}[!t]
  \centering
  \includegraphics[width=0.95\columnwidth]{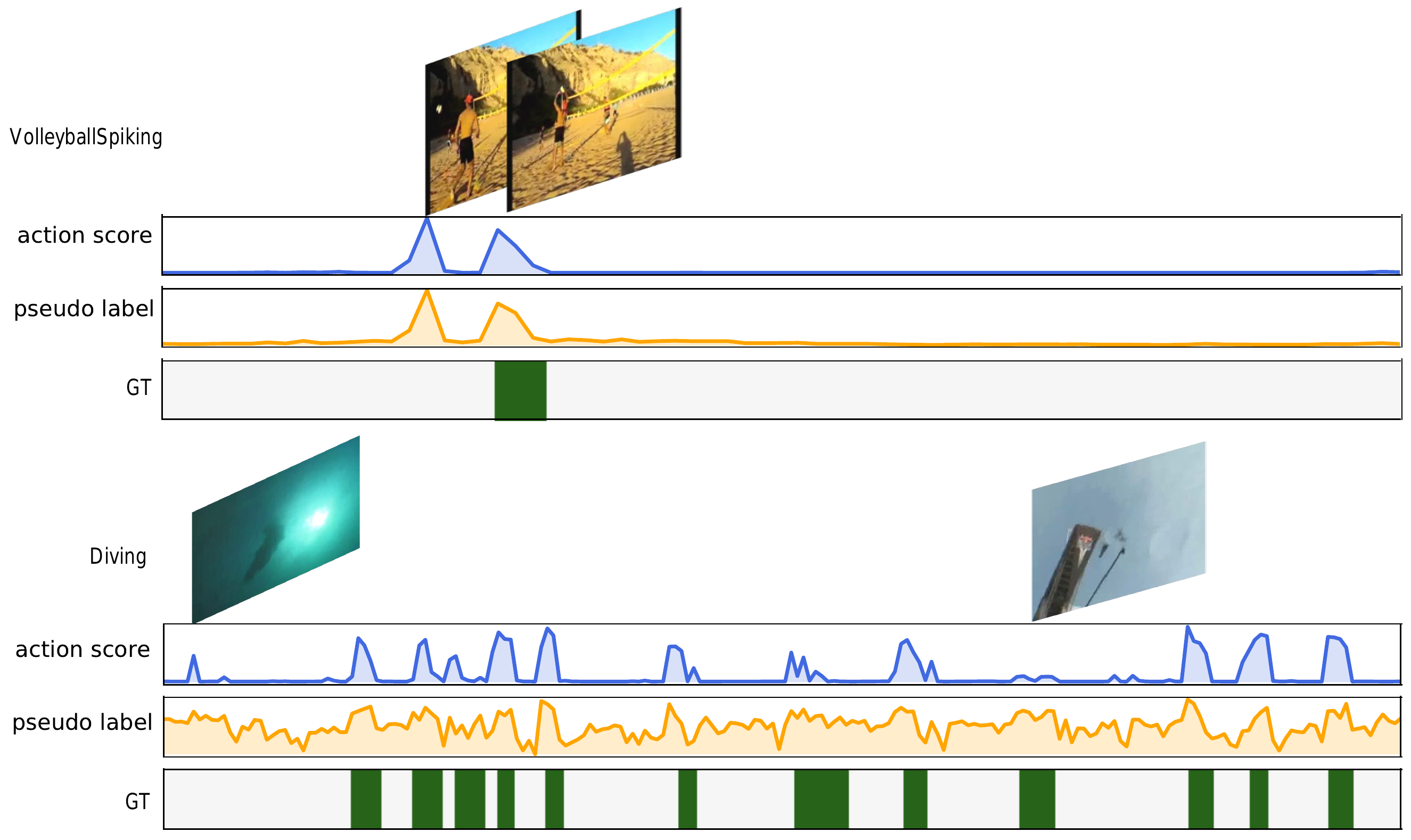}
  \caption{Some failure cases on THUMOS14 dataset. The two examples are \emph{Volleyball Spiking} and \emph{Cliff Diving}, respectively.}\label{fig:failure_case}
\vspace{-5mm}
\end{figure}

\vspace{-5mm}
\paragraph{Failure cases.}
As shown in Figure \ref{fig:failure_case}, the first failure case of \emph{Volleyball Spiking} is because the model confuses the preparing stage of this action, where the actors throw the volleyball before spiking. Due to the lack of frame-wise annotations, it is difficult to distinguish such fine-grained differences. The second failure case of \emph{Cliff Diving} comes from the clustered background and small objects. As we can see, for the first false positive instance, there is a preview segment where an actor is diving into the water. It is hard to identify those snippets without strong supervisions. For the undetected instance, the actor is too small to provide enough appearance and motion cues.
\begin{figure}[!t]
  \centering
  \includegraphics[width=0.95\columnwidth]{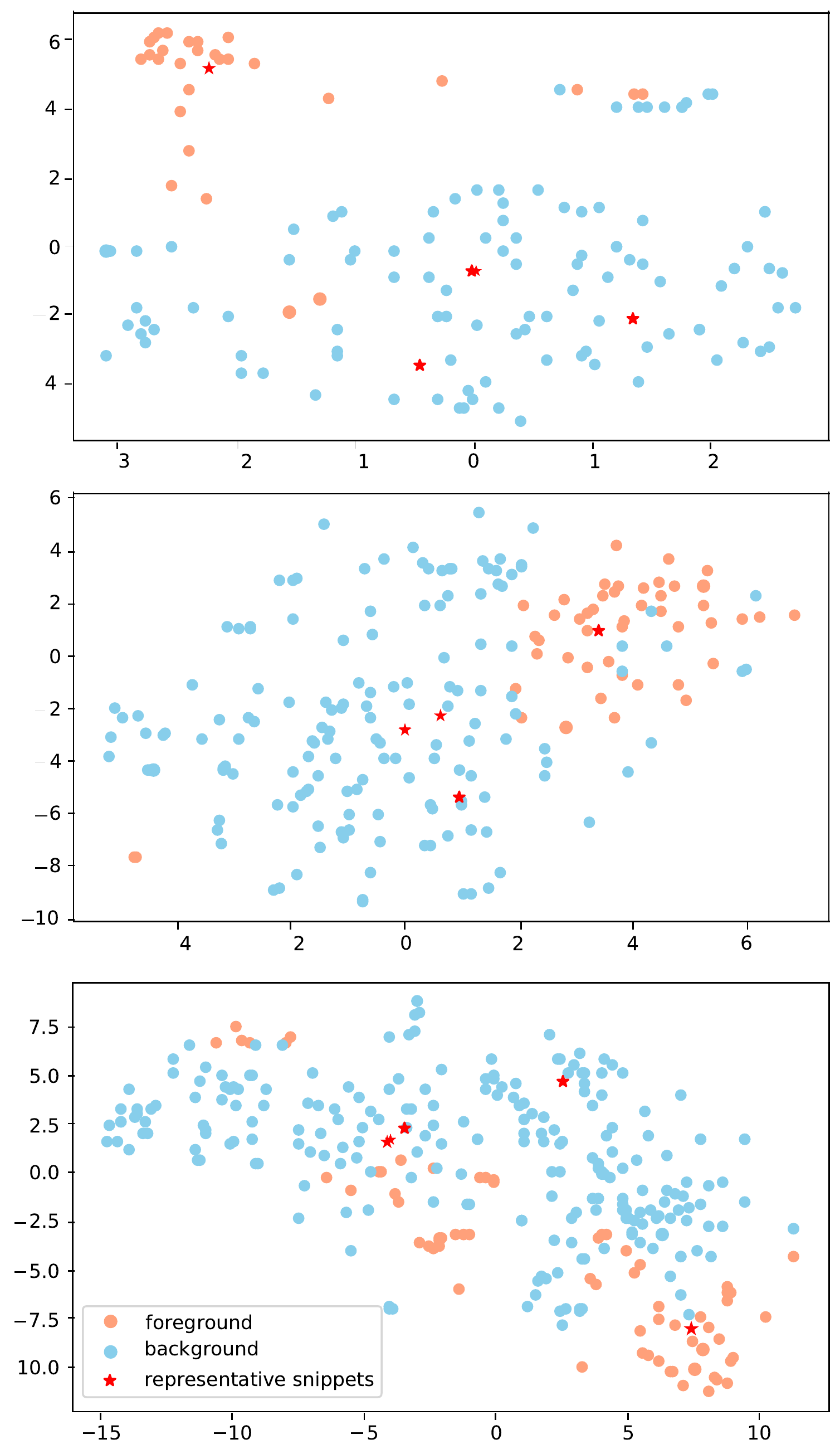}
  \caption{Visualization of the online representative snippets and the video features of the corresponding videos. The examples from top to bottom are \emph{GolfSwing}, \emph{CricketShot} and \emph{Diving}, respectively.}\label{fig:intra_visualization}
\vspace{-5mm}
\end{figure}

\vspace{-5mm}
\paragraph{Feature visualization.}
To attain further insights into the learned representative snippets, we visualize the online and offline representatives snippets and video features.
\begin{figure}[!t]
  \centering
  \includegraphics[width=0.95\columnwidth]{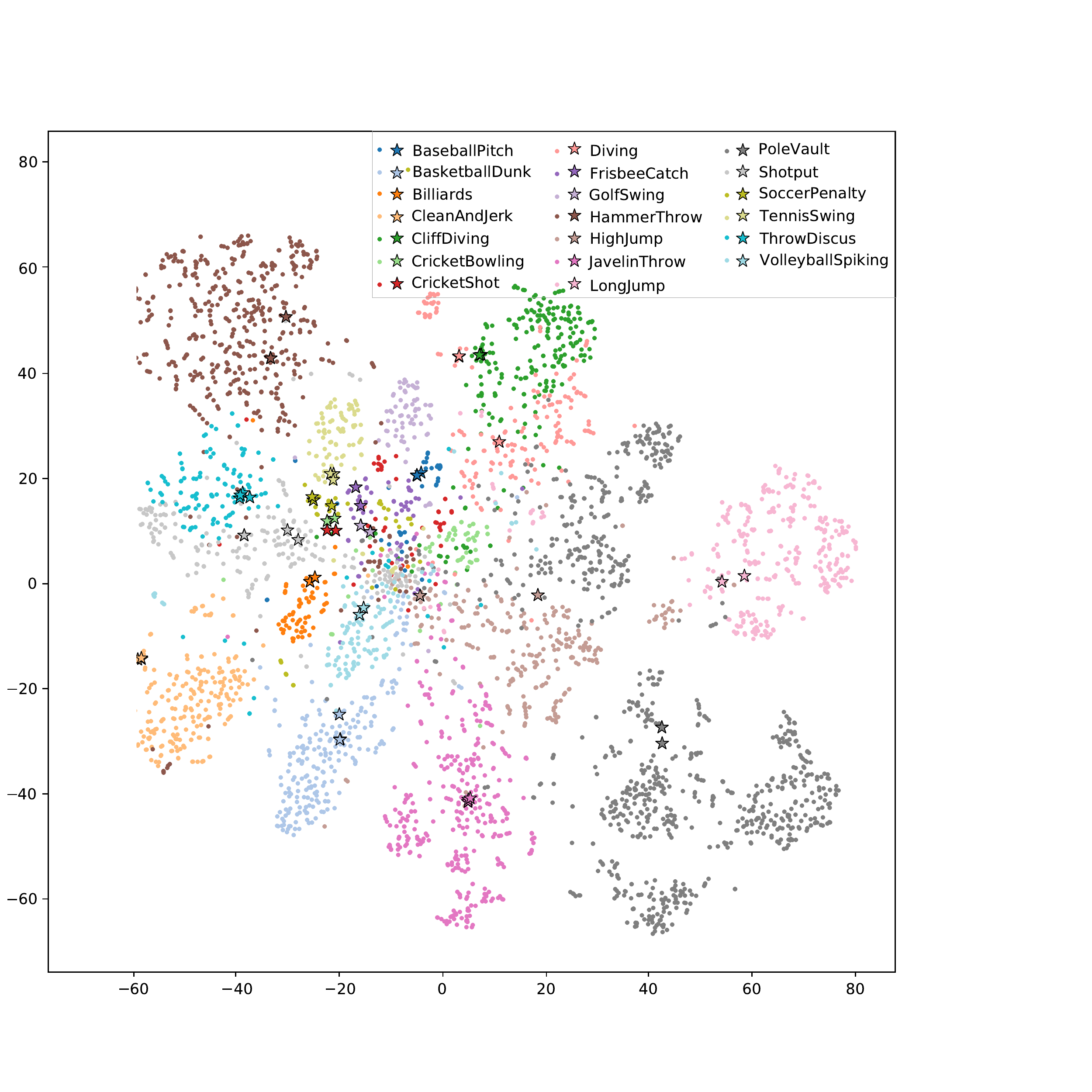}
  \caption{Visualization of the offline representative snippets and the video features of the dataset. Dots are the video features and stars are the offline representative snippets.}\label{fig:inter_visualization}
\vspace{-5mm}
\end{figure}

In Figure \ref{fig:intra_visualization}, we visualize the online representatives snippets and the video features of the corresponding videos. As we can see, the online representative snippets can well model the variations of the video, so as to describe most of the snippets of the same class.

In Figure \ref{fig:inter_visualization}, we visualize the offline representatives snippets and the video features of the dataset. As we can see, the offline representative snippets usually locate at the dense area, demonstrating that they can represent the corresponding action class.

%%%%%%%%% REFERENCES
{\small
\bibliographystyle{ieee_fullname}
\bibliography{egbib}
}